\documentclass{article} %
\PassOptionsToPackage{numbers}{natbib}
\usepackage{iclr2024_conference,times}
\pdfoutput=1

\usepackage{amsmath,amsfonts,bm}

\def\eqref#1{equation~\ref{#1}}

\def\1{\bm{1}}

\DeclareMathAlphabet{\mathsfit}{\encodingdefault}{\sfdefault}{m}{sl}
\SetMathAlphabet{\mathsfit}{bold}{\encodingdefault}{\sfdefault}{bx}{n}

\usepackage{hyperref}
\hypersetup{colorlinks=true,linkcolor=red,citecolor=green}
\usepackage{url}

\usepackage[utf8]{inputenc} %
\usepackage[T1]{fontenc}    %
\usepackage{hyperref}       %
\usepackage{url}            %
\usepackage{booktabs}       %
\usepackage{amsfonts}       %
\usepackage{nicefrac}       %
\usepackage{microtype}      %
\usepackage{xcolor}         %
\usepackage{caption}

\renewcommand{\cite}[1]{\citep{#1}}
\newcommand{\revise}[1]{{#1}}

\usepackage{graphicx}

\usepackage{listings}
\usepackage{changepage}

\usepackage{amsmath}

\usepackage{enumitem}

\usepackage{wrapfig}

\usepackage{multirow} 

\newcommand{\ie}{\textit{i.e.}}
\newcommand{\eg}{\textit{e.g.}}
\newcommand{\etal}{\textit{et al.}}

\newcommand{\fsenc}{f^{enc}_s}
\newcommand{\fsproj}{f^{proj}_s}
\newcommand{\fvis}{f^{vis}}
\newcommand{\senc}{s^{enc}}

\newcommand{\fsdec}{f^{dec}_s}
\newcommand{\sdec}{\hat{s}^{dec}}

\newcommand{\fdenc}{f^{enc}_d}
\newcommand{\flang}{f^{lang}}
\newcommand{\fdproj}{f^{proj}_d}
\newcommand{\denc}{d^{enc}}

\newcommand{\fadec}{f^{dec}_a}
\newcommand{\adec}{\hat{a}^{dec}}
\newcommand{\facls}{f^{cls}_a}

\newcommand{\lsenc}{L^{enc}_s}
\newcommand{\lsdec}{L^{dec}_s}
\newcommand{\ladec}{L^{dec}_a}

\title{SCHEMA: State CHangEs MAtter for \\
Procedure Planning in Instructional Videos}

\author{Yulei Niu$^1$
\quad Wenliang Guo$^1$ \quad Long Chen$^2$ \quad Xudong Lin$^1$ \quad Shih-Fu Chang$^1$ \\
$^1$Columbia University \quad $^2$The Hong Kong University of Science and Technology\\
\texttt{yn.yuleiniu@gmail.com} \\
}

\iclrfinalcopy %
\begin{document}

\maketitle

\begin{abstract}

We study the problem of procedure planning in instructional videos, which aims to make a goal-oriented sequence of action steps given partial visual state observations. The motivation of this problem is to learn a \textit{structured and plannable state and action space}. Recent works succeeded in sequence modeling of \textit{steps} with only sequence-level annotations accessible during training, which overlooked the roles of \textit{states} in the procedures. In this work, we point out that State CHangEs MAtter (SCHEMA) for procedure planning in instructional videos. We aim to establish a more structured state space by investigating the causal relations between steps and states in procedures. Specifically, we explicitly represent each step as state changes and track the state changes in procedures. For step representation, we leveraged the commonsense knowledge in large language models (LLMs) to describe the state changes of steps via our designed chain-of-thought prompting. For state change tracking, we align visual state observations with language state descriptions via cross-modal contrastive learning, and explicitly model the intermediate states of the procedure using LLM-generated state descriptions. Experiments on CrossTask, COIN, and NIV benchmark datasets demonstrate that our proposed SCHEMA model achieves state-of-the-art performance and obtains explainable visualizations.

\end{abstract}

\section{Introduction}\label{sec:1}
Humans are natural experts in procedure planning, \ie, arranging a sequence of instruction steps to achieve a specific goal. Procedure planning is an essential and fundamental reasoning ability for embodied AI systems and is crucial in complicated real-world problems like robotic navigation~\cite{tellex2011understanding,jansen2020visually,brohan2022can}. Instruction steps in procedural tasks are commonly state-modifying actions that induce \textit{state changes} of objects. For example, for the task of ``\texttt{grilling steak}'', a \textit{raw} steak would be first \textit{topped with pepper} after ``\texttt{seasoning the steak}'', then \textit{placed on the grill} before ``\texttt{closing the lid}'', and become \textit{cooked pieces} after ``\texttt{cutting the steak}''. These before-states and after-states reflect fine-grained information like shape, color, size, and location of entities.
Therefore, the planning agents need to figure out both the temporal relations between action steps and the causal relations between steps and states.

\begin{figure}
\includegraphics[width=0.98\textwidth]{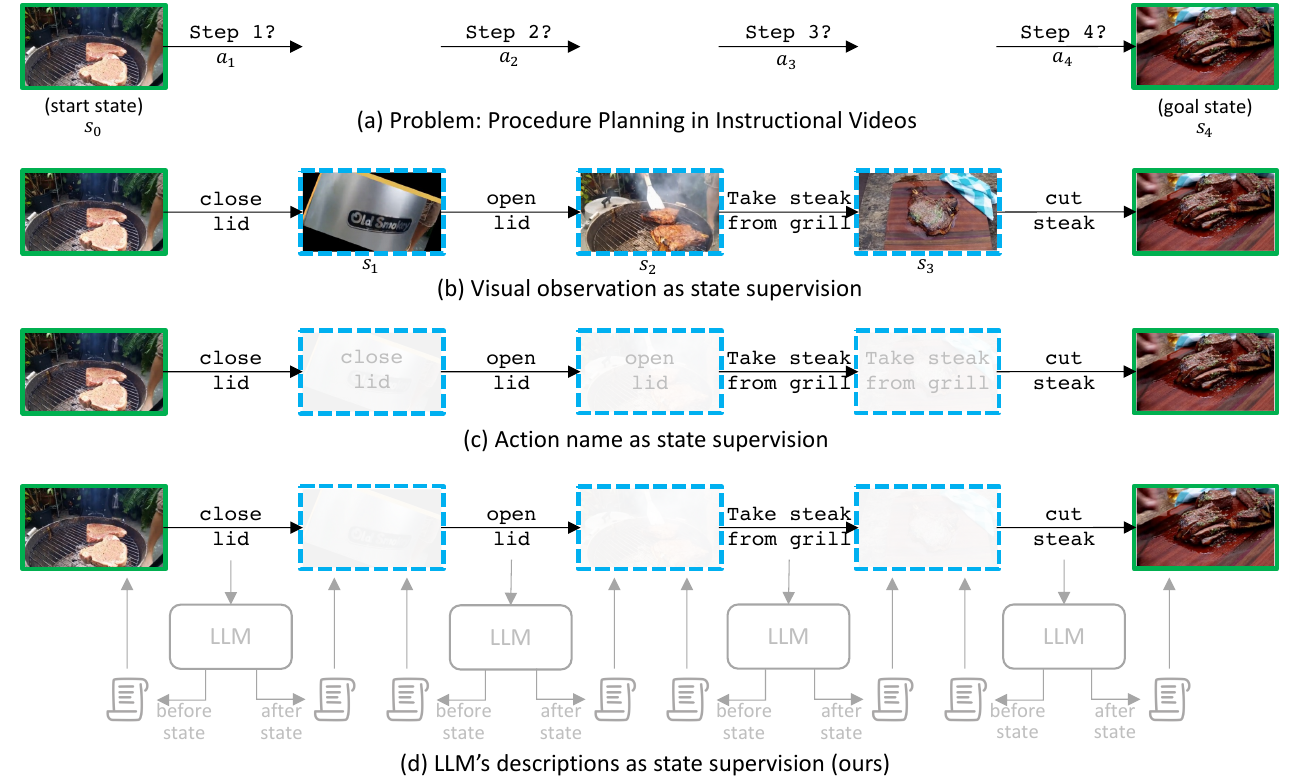}
\vspace{-0.5em}
\captionof{figure}{(a) The procedure planning is to predict a sequence of action steps given the visual observations of start and goal states. (b) For the full supervision setting, intermediate states are annotated with timestamps in the videos. (c) For the weak supervision setting, only step names are annotated and taken as supervisions of intermediate states. (d) We represent each step as state changes and take LLMs-generated descriptions for state representation learning.}
\label{fig:task}
\vspace{-4mm}
\end{figure}

Instructional videos are natural resources for learning procedural activities from daily tasks. \citet{chang2020procedure} proposed the problem of procedure planning in instructional videos, which is to produce a sequence of action steps given the visual observations of start and goal states, as shown in Figure~\ref{fig:task} \textcolor{red}{(a)}. The motivation for this problem is to \textit{learn a structured and plannable state and action space}~\cite{chang2020procedure}. While earlier works~\cite{chang2020procedure,sun2022plate,bi2021procedure} utilized the full annotations of step sequences and intermediate states as supervision (Figure~\ref{fig:task}\textcolor{red}{(b)}), recent works~\cite{zhao2022p3iv,wang2023pdpp} achieved promising results with weaker supervision, where only step sequence annotations are available during training (Figure~\ref{fig:task}\textcolor{red}{(c)}). The weaker supervision setting reduces the expensive cost of video annotations and verifies the necessity of \textit{plannable action space}. 
However, as the intermediate visual states are excluded during both training and evaluation, how to comprehensively represent the state information remains an open question.

This paper aims to establish a more \textit{structured state space} for procedure planning by investigating the causal relations between steps and states in procedures. We first ask: how do humans recognize and understand and distinguish steps in procedures? Instead of solely focusing on the action information, humans would track state changes in the procedures by leveraging their commonsense knowledge, which is more informative than only looking at the actions. For example, the steak is \textit{cooked and as a whole} before ``\texttt{cutting}'' for ``\texttt{grilling steak}'', and becomes \textit{cooked pieces} after the step. Previous NLP studies have demonstrated the helpfulness of state change modeling in various reasoning tasks, including automatic execution of biology experiments~\cite{mysore2019materials}, cooking recipes~\cite{bollini2013interpreting}, and daily activities~\cite{yang2015leveraging}. Recent studies further explicitly track state changes of entities in procedure texts~\cite{mishra2018tracking,tandon2018reasoning,tandon2020dataset,zhang2023openpi2,wu2023openpi,li2023understand}. The success of state changes modeling motivates us to investigate the causal relations between steps and states for procedure planning.
 
In this work, we achieve this goal by representing each state-modifying step as state changes.
The cores of our method are step representation and state change tracking. For step representation, motivated by the success of large language models (LLMs) in visual recognition~\cite{menon2022visual}, we leveraged LLMs to describe the state changes of each step. Specifically, we asked LLMs (e.g., GPT-3.5) to describe the states before and after each step with our designed chain-of-thought prompts (Sec.~\ref{sec:3.2}). For state changes tracking, as shown in Figure~\ref{fig:task}\textcolor{red}{(d)}, we align the visual state observations with language state descriptions via cross-modal contrastive learning. Intuitively, the start visual state should be aligned with the before-state descriptions of the first step, while the goal visual state should be aligned with the after-state descriptions of the last step. As the language descriptions are more discriminative than visual states, we expect the multi-modal alignment to learn a more structured state space. We also take state descriptions as supervisions of intermediate visual states. Finally, the step prediction model is trained in a masked token prediction manner. 

Our main contributions are summarized as follows:
\begin{itemize}[leftmargin=*]
\vspace{-2mm}
\item We pointed out that State CHangEs MAtter (SCHEMA) for procedure planning in instructional videos, and
proposed a new representation of steps in procedural videos as state changes.
\item We proposed to track state changes by aligning visual state observations with LLMs-generated language descriptions for a more structured state space and represent mid-states via descriptions.
\item Our extensive experiments on CrossTask, COIN, and NIV datasets demonstrated the quality of state description generation and the effectiveness of our method.
\end{itemize}

\section{Related Work}

\noindent \textbf{Procedure Planning}~\cite{chang2020procedure,zhang2020reasoning} is an essential and fundamental problem for embodied agents.
In this work, we followed Chang \etal's~\cite{chang2020procedure} formulation of procedure planning. Recent works proposed different approaches for sequence generation, \eg, auto-regressive Transformers~\cite{sun2022plate}, policy learning~\cite{bi2021procedure}, probabilistic modeling~\cite{bi2021procedure}, and diffusion models~\cite{wang2023pdpp}. Interestingly, \citet{zhao2022p3iv} used only language instructions as supervision for procedures and did not require full annotations of intermediate visual states, which highlights the importance of sequence generation for procedure planning. 
These methods commonly formulated the problem of procedure planning as conditional sequence generation, and the visual observations of states are treated as conditional inputs. 
However, the motivation of procedure planning is to align the state-modifying actions with their associated state changes, and expect the agents to understand how the state changes given the actions. 

\noindent \textbf{Instructional Videos Analysis}.
Instructional videos have been a good data source to obtain data for procedural activities~\cite{rohrbach15ijcv,breakfast, youcook2, zhukov2019cross, tang2019coin,miech2019howto100m}.
Existing research on this topic usually tackles understanding the step-task structures in instructional videos, where the step/task annotation can be either obtained from manual annotation on a relatively small set of videos~\cite{youcook2, zhukov2019cross, tang2019coin} or through weak/distant supervision on large unlabeled video data~\cite{miech2019howto100m,miech2020end,dvornik2023stepformer,lin2022learning}. For example, StepFormer~\cite{dvornik2023stepformer} is a transformer decoder trained with video subtitles (mostly from automatic speech recognition) for discovering and localizing steps in instructional videos. 
Another recent research~\cite{souvcek2022look} tackles a more fine-grained understanding of instructional videos, which learns to identify state-modifying actions via self-supervision. However, such approaches require training on the large collection of noisy unlabeled videos, which is expensive and inaccurate enough~\cite{souvcek2022look}. 

\noindent \textbf{Tracking State Changes} is an essential reasoning ability in complex tasks like question answering and planning. Recent work has made continuous progress on explicitly tracking entity state changes in procedural texts~\cite{mishra2018tracking,tandon2018reasoning,tandon2020dataset,zhang2023openpi2,wu2023openpi,li2023understand}. Some work in the CV area also investigated the relations between actions and states in videos~\cite{alayrac2017joint,souvcek2022look,souvcek2022multi,nishimura2021state,shirai2022visual,xue2023learning,souvcek2023genhowto,saini2023chop}. 
\revise{Especially, \citet{nishimura2021state} focused on the video procedural captioning task and proposed to model material state transition from visual observation, which introduces a visual simulator modified from a natural language understanding simulator. \citet{shirai2022visual} established a multimodal dataset for object state change prediction, which consists of image pairs as state changes and workflow of receipt text as an action graph. However, the object category is limited to food or tools for cooking.} Considering the similarity between procedural texts and instructional videos it is natural to explore state changes in instructional videos. In this work, we investigate state changes in procedural videos for procedure planning.

\section{SCHEMA: \underline{S}tate \underline{CH}ang\underline{E}s \underline{MA}tter}

In this section, we introduce the details of our proposed framework, State CHangEs MAtter (SCHEMA). We first introduce the background of procedure planning in Sec.~\ref{sec:3.1}, and present our method in Sec.~\ref{sec:3.2}$\sim$\ref{sec:3.4}. Specifically, we first provide the details of step representation in Sec.~\ref{sec:3.2}, model architecture in Sec.~\ref{sec:3.3}, and training and inference in Sec.~\ref{sec:3.4}.

\subsection{Problem Formulation}\label{sec:3.1}

We follow \citet{chang2020procedure}'s formulation of procedure planning in instructional videos. As shown in Figure~\ref{fig:task}\textcolor{red}{(a)}, given the visual observations of start state $s_0$ and goal state $s_T$, the task is to plan a procedure, \ie, a sequence of action steps $\hat{\pi}=a_{1:T}$, which can transform the state from $s_0$ to $s_T$. The procedure planning problem can be formulated as $p(a_{1:T}|s_0,s_T)$. 

The motivation of this task is to learn a \textit{structured and plannable state and
action space}. For the training supervision, earlier works
used full annotations of procedures including both action steps $a_{1:T}$ and their associated visual states, \ie, the states before and after the step, which are annotated as timestamps of videos (Figure~\ref{fig:task}\textcolor{red}{(b)}). \citet{zhao2022p3iv} proposed to use weaker supervision where only the action step annotations $a_{1:T}$ are available (Figure~\ref{fig:task}\textcolor{red}{(c)}), which reduces the expensive annotation cost for videos. Recent works under this setting show that the \textit{plannable action space} can be established by conditional sequence generation~\cite{zhao2022p3iv,wang2023pdpp}. However, there remain open questions about the role of \textit{structured state space}: why are intermediate states optional for procedure planning? Are there any better representations for steps and visual states?

\subsection{Step Representation As State Changes In Language}\label{sec:3.2}

Our goal is to construct a more \textit{structured state space} by investigating the causal relations between steps and states in procedures. Motivated by the success of state changes modeling in various reasoning tasks~\cite{bollini2013interpreting,yang2015leveraging,mysore2019materials}, we represent steps as their before-states and after-states. The state changes can be represented by visual observations or language descriptions. 
We observed that visual scenes in instructional videos are diverse and noisy, and the details are hard to capture if the object is far from the camera. 
In addition, the intermediate visual states may not be available due to the high cost of video annotations. Therefore, we represent state changes as discriminative and discrete language descriptions.

Motivated by \citet{menon2022visual}, we leveraged large language models (LLMs), such as GPT-3.5~\cite{brown2020language}, to generate language descriptions of states based on their commonsense knowledge. In short, we fed each action step with its high-level task goal to the LLMs, and query several descriptions about the associate states before and after the action step. 
A baseline prompting following~\citet{menon2022visual} for state descriptions is:
\begin{adjustwidth}{1cm}{1cm}
\begin{lstlisting}[breakatwhitespace=true]
Q: What are useful features for distinguishing the states before and after [step] for [goal] in a frame?
A: There are several useful visual features to tell the state changes before and after [step] for [goal]:
\end{lstlisting}
\end{adjustwidth}
\vspace{-2mm}
However, we empirically found that this prompting does not work well for state descriptions, as LLMs disregard the commonsense knowledge behind the step. For example, given the action step ``\texttt{add onion}'' and the task ``\texttt{make kimchi fried rice}'', the before-state description is ``the onion was uncut and unchopped'', which is incorrect because the onion should have been cut.

\begin{figure}
\centering
\includegraphics[width=1.0\textwidth]{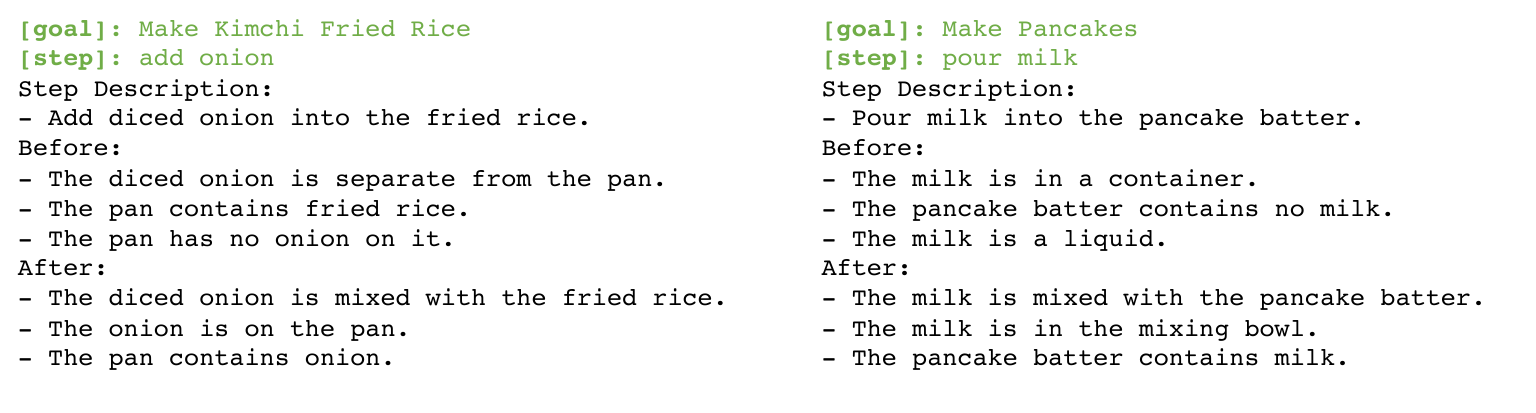}
\caption{Examples of GPT-3.5 generated descriptions using our chain-of-thought prompting.}
\label{fig:desc}
\end{figure}

\noindent\textbf{Chain-of-thought Prompting}. Aware of the misalignment between action and states, we proposed a chain-of-thought prompting~\cite{wei2022chain} strategy to first describe the details of action steps and then describe the state changes according to the details of the steps. Our prompt is designed as:
\begin{adjustwidth}{1cm}{1cm}
\begin{lstlisting}[breakatwhitespace=true]
First, describe the details of [step] for [goal] with one verb. 
Second, use 3 sentences to describe the status changes of objects before and after [step], avoiding using [verb].
\end{lstlisting}
\end{adjustwidth}
\vspace{-2mm}
where ``\texttt{[verb]}'' is the action name (\eg, ``pour'') to increase the description diversity. We also provide several examples as context (see appendix). We fixed the number of descriptions as 3 as we empirically found that one or two descriptions cannot cover all the objects and attributes, while more than three descriptions are redundant. Figure~\ref{fig:desc} illustrates two examples of the generated descriptions based on our prompts. Overall, the step and state descriptions contain more details about attributes, locations, and relations of objects. 
In the following, for the step name $A_i$, we denote its step description as $d^s_i$, before-state descriptions as $\{d^b_{i1}, \cdots, d^b_{iK}\}$, and after-state descriptions as $\{d^a_{i1}, \cdots, d^a_{iK}\}$, where $K\!=\!3$ in our implementation.

\subsection{Architecture}\label{sec:3.3}

\begin{figure}
\centering
\includegraphics[width=0.98\textwidth]{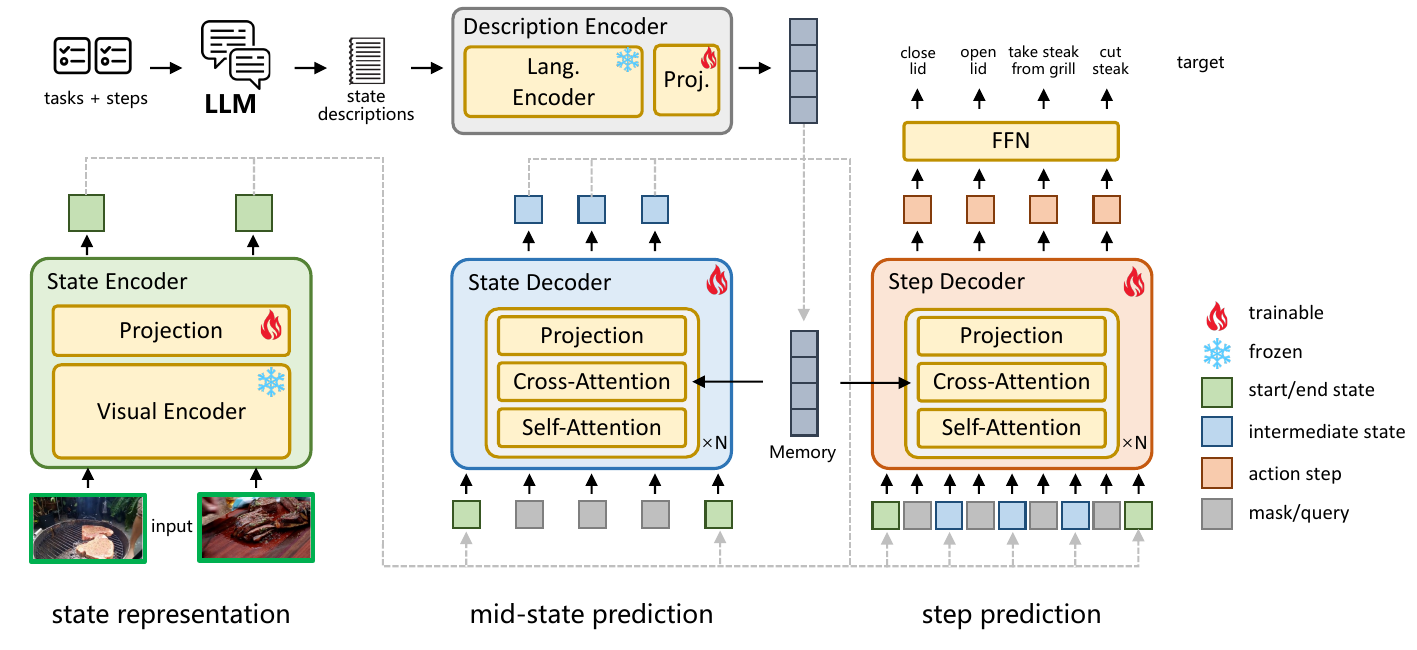}
\vspace{-3mm}
\caption{The pipeline of our SCHEMA for procedure planning.}
\label{fig:pipeline}
\vspace{-4mm}
\end{figure}

Figure~\ref{fig:pipeline} illustrates the overview of our SCHEMA pipeline. Overall, we break up the procedure planning problem $p(a_{1:T}|s_0,s_T)$ into two subproblems, \ie, mid-state prediction and step prediction. Mid-state prediction is to estimate the intermediate states $s_{1:(T-1)}$ given $s_0$ and $s_T$, \ie, $p(s_{1:(T-1)}|s_0,s_T)$. Step prediction is to predict the step sequence given the full states, \ie, $p(a_{1:T}|s_{0:T})$. We formulate the procedure planning problem as:
\begin{equation}
    p(a_{1:T}|s_0,s_T)=\int \underbrace{p(a_{1:T}|s_{0:T})}_{\text{step prediction}}
    \underbrace{p(s_{1:(T-1)}| s_0, s_T)}_{\text{mid-state prediction}} ds_{1:(T-1)}.
\end{equation}

\subsubsection{State Representation}

We align visual observations with language descriptions of the same states to learn a structure state space, which will be introduced in Sec.~\ref{sec:3.4}.

\textbf{State encoder}. The state encoder takes the video frame as input and outputs its embedding. The state encoder $\fsenc$ consists of a fixed pre-trained visual feature extractor $\fvis$ and a trainable projection (two-layer FFN) $\fsproj$. The embedding for state $s$ is obtained by $\senc=\fsenc(s)=\fsproj(\fvis(s))$. 

\textbf{Description encoder}. Similar to the state encoder, the description encoder $\fdenc$ consists of a fixed language feature extractor $\flang$ and a trainable projection $\fdproj$. The description encoder takes description $d$ as input and outputs its embedding $\denc=\fdenc(d)=\fdproj(\flang(d))$.

\subsubsection{Mid-state Prediction}\label{sec:3.3.2}

\noindent\textbf{State Decoder}. The state decoder $\fsdec$ is an non-autoregressive Transformer~\cite{vaswani2017attention}. 
The state decoder predicts the intermediate states $s_{1:(T-1)}$ given the start state $s_0$ and the goal state $s_T$. The query for the state decoder is denoted as $Q_{s}=[\senc_0+p_0, p_1, \cdots, p_{T-1}, \senc_T+p_T]$, where $p_i$ denotes the $i$-th positional embedding. The state decoder also takes the collection of state descriptions $D_s=\{d^b_{11}, \cdots, d^b_{CK}, d^a_{11}, \cdots, d^a_{CK}\}$ as the external memory $M$, where $C$ is the number of step classes and $M=\fdenc(D_s)$. The external memory interacts with the decoder via cross-attention. The state decoding process to obtain the embeddings $\sdec_i$ is denoted as:
\begin{equation}\label{eq:fsdec}
    \sdec_1, \cdots, \sdec_{T-1} = \fsdec(Q_{s}, M).
\end{equation}

\subsubsection{Step Prediction}\label{sec:3.3.3}
\noindent\textbf{Step Decoder}. The step decoder $\fadec$ is a Transformer model with a similar architecture as the state decoder $\fsdec$. The query combines state embeddings and positional embeddings, denoted as $Q_a=[\senc_0+q_0,q_1,\sdec_1+p_2,\cdots,\sdec_{T-1}+q_{2T-2},q_{2T-1},\senc_T+q_{2T}]$
where $q_i$ denotes the $i$-th positional embedding. Similar to the state decoder $\fsdec$, the step decoder $\fadec$ also takes $M=\fdenc(D_s)$ as the external memory. The step decoding process is denoted as:
\begin{equation}\label{eq:fadec}
    \adec_1, \cdots, \adec_{T} = \fadec(Q_{a}, M),
\end{equation}
where $\adec_1, \cdots, \adec_{T}$ are the estimated action embeddings. A two-layer feed-forward network (FFN) $\facls$ is built on top of $\adec$ as the step classifier to predict the logits of steps, \ie, $\hat{a}=\facls(\adec)$.

In addition to capturing the task information, we establish a task classifier that takes the visual features of start and end states as input and outputs a vector to represent the task information~\cite{wang2023pdpp,wang2023event}. The task feature is added to the queries $Q_s$ and $Q_a$, where we omitted it for simplicity.

\subsection{Training and Inference}\label{sec:3.4}

The training process consists of three parts: (1) state space learning that aligns visual observations with language descriptions, (2) masked state modeling for mid-state prediction, and (3) masked step modeling for step prediction. For simplicity, we define the losses with one procedure.

\begin{figure}
\centering
\includegraphics[width=1.0\textwidth]{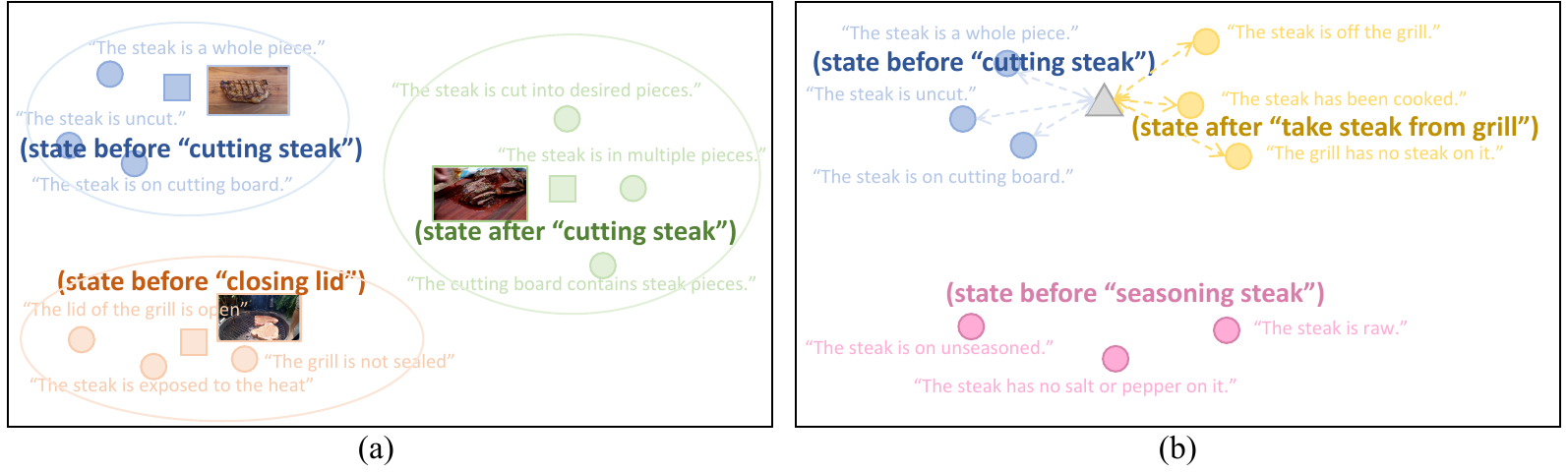}
\vspace{-3mm}
\caption{(a) For structured state space learning, we force the visual state observation to be close to its language descriptions and away from others. (b) For mid-state prediction, we use the before-state descriptions and after-state descriptions as guidance to learn the presentation of mid-states.}
\label{fig:state}
\end{figure}

\noindent\textbf{State Space Learning}. Although vision-language models like CLIP~\cite{radford2021learning} are pre-trained for vision-language alignment, they cannot be directly used for our problem as the pre-training is not designed for fine-grained state understanding. Therefore, we train two additional projections $\fsproj$ and $\fdproj$ on top of the visual encoder and language encoder. The added projections also allow us to align other visual features with language features. Given the start state $s_0$ (or end state $s_T$) and a step label $a_i$, the similarity between $s_0$ (or $s_T$) and \revise{each step} $A_i$ is calculated by $sim(s_0, A_i)\!=\! \sum_{j=1}^K\!<\!\senc_0,d^{enc,b}_{ij}\!>$ and $sim(s_T, A_i)\!=\! \sum_{j=1}^K\!<\!\senc_T,d^{enc,a}_{ij}\!>$,
where $<\cdot,\cdot>$ denotes the dot product. Figure~\ref{fig:state} \textcolor{red}{(a)} illustrates the idea of structured state space via vision-language alignment. Specifically, we regard the language descriptions with the same state as positive samples, and take descriptions of \revise{the other states} as negative samples. We define the contrastive loss as:
\begin{equation}\label{eq:lsenc}
    \lsenc=\underbrace{-\log\frac{\exp(sim(s_0,A_{a_1}))}{\sum^C_{i=1}\exp(sim(s_0,A_i))}}_{\text{start state}} \underbrace{- \log\frac{\exp(sim(s_T,A_{a_T}))}{\sum^C_{i=1}\exp(sim(s_T,A_i))}}_{\text{end state}}
\end{equation}

\noindent\textbf{Masked State Modeling}. The mid-state prediction process can be regarded as a masked state modeling problem, where the intermediate states are masked from the state sequence, and the state decoder recovers the masked states. Since the annotations of intermediate states in videos are not available, we instead use LLMs-generated state descriptions as guidance. In a procedure $(s_0, a_1, s_1, \cdots, s_{T-1},a_{T}, s_T)$ where the mid-state $s_t$ is the after-state for action $a_t$ and before-state for action $a_{t+1}$, we use the before-state descriptions of $a_t$ and after-state descriptions of $a_{t+1}$ as the supervision for $s_t$. Figure~\ref{fig:state}\textcolor{red}{(b)} illustrates the idea of mid-state learning. We average the description embeddings as the target embedding $\sdec_t$ for $s_t$, and calculate the mean squared error between $\sdec_t$ and $s^{dec}_t$ as the mid-state prediction loss:
\begin{equation}
    s^{dec}_t = \frac{1}{2K}(\sum^K_{j=1}d^{enc,a}_{a_{t,j}} + d^{enc,b}_{a_{{t+1},j}}),\qquad
    L^{dec}_s=\sum^{T-1}_{t=1}(\sdec_t-s^{dec}_t)^2.
\end{equation}

\noindent\textbf{Masked Step Modeling}. Similar to mid-state estimation, the step prediction process can also be regarded as a masked step modeling problem, where the steps are masked from the state-action sequences. The loss is the cross-entropy between ground-truth answers $a_t$ and predictions $\adec_t$, \ie, $L^{dec}_a=\sum_{t=1}^T-a_t\log\adec_t$. The final loss combines the above losses, \ie, $L=\lsenc+\lsdec+\ladec$.

\textbf{Inference}. The non-autoregressive Transformer model may make insufficient use of the temporal relation information among the action steps, \ie, action co-occurrence frequencies. Inspired by the success of Viterbi algorithm\cite{viterbi1967error} in sequential labeling works~\cite{koller2016deep,koller2017re,richard2017weakly,richard2018neuralnetwork}, we follow \citet{zhao2022p3iv} and conduct the Viterbi algorithm for post-processing during inference. For Viterbi, we obtained the emission matrix based on the probability distribution over $[\adec_1,\cdots,\adec_T]$, and estimated the transition matrix based on action co-occurrence frequencies in the training set. Different from \citet{zhao2022p3iv} that applied Viterbi to probabilistic modeling, we applied Viterbi to deterministic modeling. Specifically, instead of sampling 1,500 generated sequences to estimate the emission matrix~\cite{zhao2022p3iv}, we run the feedforwarding only once and use the single predicted probability matrix as the emission matrix, which is simpler and more efficient.

\section{Experiments}\label{sec:}

\begin{table}[t]
\centering
\caption{Comparison with other methods on CrossTask dataset.}
\vspace{-3mm}
\scalebox{0.92}{
\begin{tabular}{lccccccccc}
\toprule
& \multicolumn{3}{c}{$T$ = 3} & & \multicolumn{3}{c}{$T$ = 4} & $T$=5 & $T$ = 6 \\ 
\cline{2-4} \cline{6-8}
{Models}           & SR$\uparrow$    & mAcc$\uparrow$   & mIoU$\uparrow$  &   & SR$\uparrow$    & mAcc$\uparrow$   & mIoU$\uparrow$  & SR$\uparrow$ & SR$\uparrow$ \\ \midrule
{Random} &   $<$0.01    &    0.94    &   1.66    &   &   $<$0.01    &    0.83    &   1.66   & $<$0.01 & $<$0.01  \\
{Retrieval-Based}   &   8.05    &   23.30     &   32.06    &   &   3.95    &    22.22    &    36.97   & 2.40 & 1.10  \\
{WLTDO}            &   1.87    &    21.64    &   31.70   &    &   0.77    &    17.92    &    26.43  & --- & ---  \\
{UAAA}             &   2.15    &   20.21     &   30.87   &    &   0.98    &   19.86     &  27.09   & --- & ---   \\
{UPN}               &   2.89    &    24.39    &   31.56     &  &   1.19    &   21.59     &   27.85   & --- & ---  \\
{DDN}              &   12.18    &    31.29    &    47.48    &  &   5.97    &    27.10    &    48.46  & 3.10 & 1.20  \\
{PlaTe}              &   16.00    &    36.17    &    65.91   &   &   14.00    &    35.29    &    55.36   & --- & --- \\
{Ext-MGAIL w/o Aug.}         &   18.01 &    43.86   & 57.16   &    &   ---   &    ---    &   ---    & --- & --- \\
{Ext-GAIL}         &   21.27    &   49.46     &   61.70    &   &   16.41    &    43.05    &   60.93   & --- & ---  \\
{P$^3$IV w/o Adv.} & 22.12 & 45.57 & 67.40 & & --- & --- & --- & --- & --- \\
{P$^3$IV}             &   23.34    &   49.96     &   73.89   &    &   13.40    &   44.16     &   70.01  & 7.21 & 4.40   \\
{EGPP}             & 26.40 & 53.02 & 74.05 & & 16.49 & 48.00 & 70.16 & 8.96 & 5.76 \\
\midrule
{SCHEMA (Ours)}             &   \bf 31.83 & \bf 57.31 & \bf 78.33  &  & \bf 19.71 & \bf 51.85 & \bf 74.46 & \bf 11.41 & \bf 7.68 \\
\bottomrule
\end{tabular}
}

\label{tab:sota}
\end{table}

\subsection{Evaluation Protocol}\label{sec:eval}

\noindent \textbf{Datasets}. We evaluate our SCHEMA method on three benchmark instruction video datasets, CrossTask~\cite{zhukov2019cross}, and COIN~\cite{tang2019coin}, and NIV~\cite{alayrac2016unsupervised}. The CrossTask dataset consists of 2,750 videos from 18 tasks depicting 133 actions, with an average of 7.6 actions per video. The COIN dataset contains 11,827 videos from 180 tasks, with an average of 3.6 actions per video. The NIV dataset contains 150 videos with an average of 9.5 actions per video. Following previous works~\cite{chang2020procedure,bi2021procedure,sun2022plate}, we randomly select 70\% of the videos in each task as the training set and take the others as the test set. 
We extract all the step sequences $a_{t:(t+T-1)}$ in the videos as procedures with the time horizon $T$ as 3 or 4.

\noindent \textbf{Feature Extractors}. As our goal for state space learning is to align visual observation and language descriptions, we tried CLIP~\cite{radford2021learning} ViT-L/14 model as visual encoder and its associated pretrained Transformer as language encoder. We also follow recent works and use the S3D network~\cite{miech2019howto100m} pretrained on the HowTo100M dataset~\cite{miech2020end} as the visual encoder, and add two projections for vision-language alignment (Sec.~\ref{sec:3.4}). We empirically found that video-based pre-trained features perform better than image-based pre-trained features. In the following, we used the pretrained S3D model as visual encoder.

\noindent \textbf{Metrics}. Following previous works~\cite{chang2020procedure,bi2021procedure,sun2022plate,zhao2022p3iv}, we evaluate the models on three metrics. (1) Success Rate (SR) is the most strict metric that regards a procedure as a success if all the predicted action steps in the procedure match the ground-truth steps. (2) mean Accuracy (mAcc) calculates the average accuracy of the predicted actions at each step. (3) mean Intersection over Union (mIoU) is the least strict metric that calculates the overlap between the predicted procedure and ground-truth plan, obtained by $\frac{|\{a_t\}\cap\{\hat{a}_t\}|}{|\{a_t\}\cup\{\hat{a}_t\}|}$, where $\{\hat{a}_t\}$ is the set of predicted actions and $\{a_t\}$ is the set of ground truths.

\textbf{Baselines}. We follow previous works and consider the following baseline methods for comparisons. The recent baselines are (1) PlaTe~\cite{sun2022plate}, which extends DNN and uses a Transformer-based architecture; (2) Ext-GAIL~\cite{bi2021procedure}, which uses reinforcement learning for procedure planning; (3) P$^3$IV~\cite{zhao2022p3iv}, which is the first to use weak supervision and proposed a generative adversarial framework; (4) PDPP~\cite{wang2023pdpp}, which is a diffusion-based model for sequence distribution modeling; and (5) EGPP~\cite{wang2023event}, which extracts event information for procedure planning. Details of other earlier baselines can be found in appendix.

\begin{table}[t]
\centering
\caption{Comparison with PDPP on CrossTask dataset. $\dagger$ denotes the results under PDPP's setting.}
\scalebox{0.92}{
\begin{tabular}{lccccccccc}
\toprule
& \multicolumn{3}{c}{$T$ = 3} & & \multicolumn{3}{c}{$T$ = 4} & $T$=5 & $T$ = 6 \\ 
\cline{2-4} \cline{6-8}
{Models}           & SR$\uparrow$    & mAcc$\uparrow$   & mIoU$\uparrow$  &   & SR$\uparrow$    & mAcc$\uparrow$   & mIoU$\uparrow$  & SR$\uparrow$ & SR$\uparrow$ \\ \midrule
{PDPP} & 26.38 & 55.62 & 59.34 & & 18.69 & \bf 52.44 & 62.38 & \bf 13.22 & 7.60 \\
{SCHEMA (Ours)} & \bf 31.83 & \bf 57.31 & \bf 78.33  &  & \bf 19.71 & 51.85 & \bf 74.46 & 11.41 & \bf 7.68 \\
\hline
{PDPP}$^\dagger$ & 37.20 & \bf 64.67 & 66.57 & & 21.48 & 57.82 & 65.13 & 13.45 & 8.41 \\
{SCHEMA (Ours)}$^\dagger$ & \bf 38.93 & 63.80 & \bf 79.82 & & \bf 24.50 & \bf 58.48 & \bf 76.48 & \bf 14.75 & \bf 10.53 \\
\bottomrule
\end{tabular}
}
\label{tab:pdpp}
\end{table}

\begin{table}[t]
\centering
\caption{Comparisons with other methods on COIN dataset.}
\scalebox{0.95}{
\begin{tabular}{l ccc c ccc}
\toprule
& \multicolumn{3}{c}{$T$ = 3} & & \multicolumn{3}{c}{$T$ = 4} \\ 
\cline{2-4} \cline{6-8}
{Models}           & SR$\uparrow$    & mAcc$\uparrow$   & mIoU$\uparrow$  &   & SR$\uparrow$    & mAcc$\uparrow$   & mIoU$\uparrow$ \\ \midrule
Random       &   <0.01    &    <0.01    &    2.47    &  &   <0.01    &    <0.01    &    2.32    \\
Retrieval       &   4.38    &    17.40    &    32.06    &  &   2.71    &    14.29    &    36.97    \\
{DDN}              &   13.90    &    20.19    &    64.78    &  &   11.13    &    17.71    &    68.06    \\
{P$^3$IV}             &   15.40    &   21.67    &   76.31   &    &   11.32    &   18.85     &   70.53   \\
{EGPP}             &   19.57   &   31.42 & \bf 84.95 & & 13.59 & 26.72 & \bf 84.72 \\
\midrule
{SCHEMA (Ours)} & \bf 32.09 & \bf 49.84 & 83.83 & & \bf 22.02 & \bf 45.33 & 83.47 \\
\bottomrule
\end{tabular}
}
\label{tab:coin}
\end{table}

\subsection{Results}\label{sec:stepcls}

\textbf{Comparisons with Other Methods}. Tables~\ref{tab:sota} and~\ref{tab:coin} show the comparisons between our method and others on CrossTask and COIN datasets. Overall, our proposed method outperforms other methods by large margins on all the datasets and metrics. Specifically, for $T=3$ on CrossTask, our method outperforms P$^3$IV by over 8\% (31.83 vs. 23.34) on the sequence-level metric SR and outperforms EGPP by over 5\% (31.83 vs. 26.40). The improvements are consistent with longer procedures (\ie, $T=4,5,6$), and other two step-level metrics mAcc and mIoU. We also found that both P$^3$IV and EGPP didn't work well on COIN compared to their performances on CrossTask. Specifically, P$^3$IV outperforms DDN by large margins on CrossTask (\eg, 23.34 vs. 12.18 for SR with $T=3$). However, the improvements on COIN become marginal, especially for longer procedures (\ie, 11.32 vs. 11.13 for SR with $T=4$). The similar results are observed on EGPP, As comparisons, the SR of our SEPP is consistently larger than P3IV by over 16\% for $T=3$ and 10\% for $T=4$. The improvements of mACC and mIoU are also significant. These results demonstrate the better generality and effectiveness of our method on different datasets compared to P3IV and EGPP.

An exception case is the recent work PDPP~\cite{wang2023pdpp} which achieves higher performance on both datasets. However, we argued that they define the start state and end state differently. Specifically, previous works define states as a 2-second window \textit{around} the start/end time, while PDPP defines the window \textit{after} the start time and \textit{before} the end time. Such a definition is more likely to access step information especially for short-term actions, leading to unfair comparisons with other methods. We further compared our method with PDPP under both conventional setting and their setting. The results on Table~\ref{tab:pdpp} show that our method outperforms PDPP with $T=3$ and $T=4$ under both settings, and the main improvement of PDPP comes from the different setting with a small $T$ (\eg, $\sim$11\% increase of SR on $T\!=\!3$). An interesting observation is that the benefits of different settings to PDPP become marginal with a larger $T$, which may be credited to their diffusion model.

\begin{table}[t]
\centering
\caption{Ablation studies on state space learning and mid-state prediction on CrossTask with $T\!=\!3$.}
\scalebox{0.95}{
\begin{tabular}{c cc ccc}
\toprule
& State Align. & Mid-state Pred. &
 SR$\uparrow$    & mAcc$\uparrow$   & mIoU$\uparrow$ \\
\midrule
(a) &  &   & 28.72 & 54.72 & 76.66 \\
(b) & & \checkmark & 29.41 & 54.92 & 77.26 \\
(c) & \checkmark & & 30.15 & 56.32 & 77.68 \\
(d) & \checkmark &  \checkmark & \bf 31.83 & \bf 57.31 & \bf 78.33 \\
\bottomrule
\end{tabular}
}

\label{tab:abstate}
\end{table}

\begin{figure}
\centering
\includegraphics[width=1.03\textwidth]{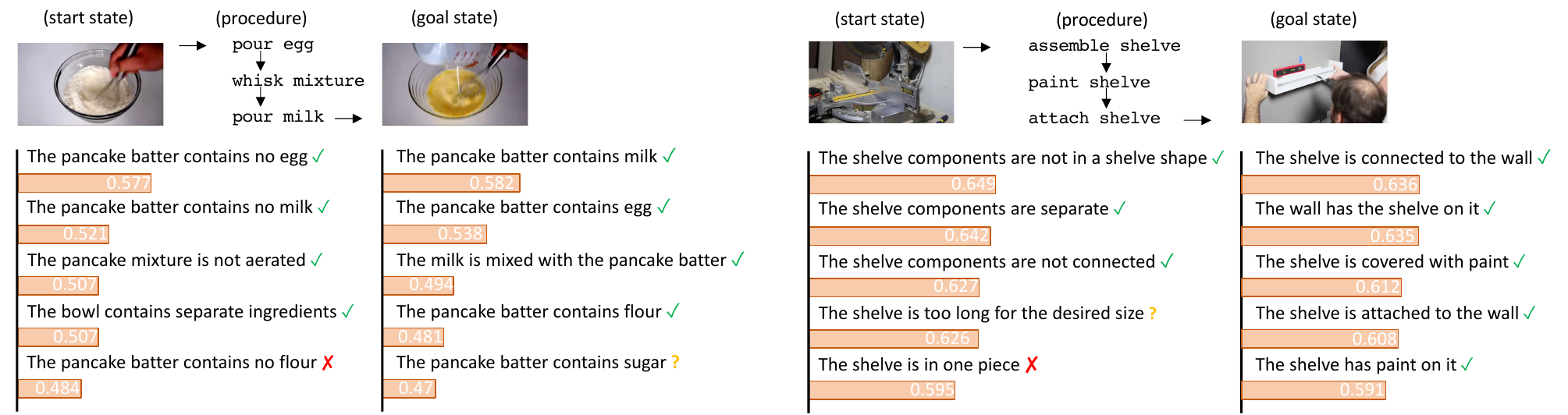}
\caption{Examples of state justifications from our model.}
\label{fig:similarity}
\end{figure}

\textbf{Ablation Studies}. We first conduct ablation studies on CrossTask to validate the effect of two key components, state alignment (Eq.~\ref{eq:lsenc} and Figure~\ref{fig:state}\textcolor{red}{(a)}) and mid-state prediction (Sec. \ref{sec:3.3.2}). As shown in Table~\ref{tab:abstate}, visual-language alignment improves SR by 1.4$\sim$2.4\% ((c) vs. (a), (d) vs. (b)) for $T=3$ on CrossTask. Also, the mid-state prediction module also improves the performance, and the improvements become larger with state alignment (\ie, +0.69\% on SR w/o state alignment vs. +1.68\% on SR with state alignment). The entire combination (d) achieves the best results. These results verified the impacts of state space learning and mid-state prediction. More ablation studies are in the appendix.

\textbf{Qualitative Results}. Figure~\ref{fig:similarity} illustrates examples of state justifications, \ie, how the model aligns visual state observation with language descriptions. We retrieve top-5 similar descriptions from the corpus of state descriptions. Overall, the retrieved descriptions match the image well, and most of the top similar descriptions are aligned with the visual observations, which improved the explainable state understanding. More visualization results are in the appendix.

\section{Conclusion}

In this work, we pointed out that State CHangEs MAtter (SCHEMA) for procedure planning in instructional videos, and proposed to represent steps as state changes and track state changes in procedural videos. We leveraged large language models (\ie, GPT-3.5) to generate descriptions of state changes, and align the visual states with language descriptions for a more structured state space. We further decompose the procedure planning into two subproblems, mid-state prediction and step prediction. Extensive experiments further verified that our proposed state representation can promote procedure planning. In the future, potential directions are to establish a benchmark dataset to explicitly track state changes in instructional videos, and to investigate the roles of state changes tracking in other procedural learning tasks like pre-training and future step forecasting. In addition, multimodal procedural planning would be a practical and challenging problem that generates coherent textual plans and visual plans, reflecting the state changes in multiple modalities. 

\section*{Acknowledgement}

This work is supported by U.S. DARPA KAIROS Program No. FA8750-19-2-1004. The views and conclusions contained herein are those of the authors and should not be interpreted as necessarily representing the official policies, either expressed or implied, of DARPA, or the U.S. Government. The U.S. Government is authorized to reproduce and distribute reprints for governmental purposes notwithstanding any copyright annotation therein.

{
\bibliographystyle{iclr2024_conference}
\bibliography{egbib}

\begin{thebibliography}{51}
\providecommand{\natexlab}[1]{#1}
\providecommand{\url}[1]{\texttt{#1}}
\expandafter\ifx\csname urlstyle\endcsname\relax
  \providecommand{\doi}[1]{doi: #1}\else
  \providecommand{\doi}{doi: \begingroup \urlstyle{rm}\Url}\fi

\bibitem[Abu~Farha \& Gall(2019)Abu~Farha and Gall]{abu2019uncertainty}
Yazan Abu~Farha and Juergen Gall.
\newblock Uncertainty-aware anticipation of activities.
\newblock In \emph{Proceedings of the IEEE/CVF International Conference on Computer Vision Workshops}, pp.\  0--0, 2019.

\bibitem[Alayrac et~al.(2016)Alayrac, Bojanowski, Agrawal, Sivic, Laptev, and Lacoste-Julien]{alayrac2016unsupervised}
Jean-Baptiste Alayrac, Piotr Bojanowski, Nishant Agrawal, Josef Sivic, Ivan Laptev, and Simon Lacoste-Julien.
\newblock Unsupervised learning from narrated instruction videos.
\newblock In \emph{Proceedings of the IEEE Conference on Computer Vision and Pattern Recognition}, pp.\  4575--4583, 2016.

\bibitem[Alayrac et~al.(2017)Alayrac, Laptev, Sivic, and Lacoste-Julien]{alayrac2017joint}
Jean-Baptiste Alayrac, Ivan Laptev, Josef Sivic, and Simon Lacoste-Julien.
\newblock Joint discovery of object states and manipulation actions.
\newblock In \emph{Proceedings of the IEEE International Conference on Computer Vision}, pp.\  2127--2136, 2017.

\bibitem[Bi et~al.(2021)Bi, Luo, and Xu]{bi2021procedure}
Jing Bi, Jiebo Luo, and Chenliang Xu.
\newblock Procedure planning in instructional videos via contextual modeling and model-based policy learning.
\newblock In \emph{Proceedings of the IEEE/CVF International Conference on Computer Vision}, pp.\  15611--15620, 2021.

\bibitem[Bollini et~al.(2013)Bollini, Tellex, Thompson, Roy, and Rus]{bollini2013interpreting}
Mario Bollini, Stefanie Tellex, Tyler Thompson, Nicholas Roy, and Daniela Rus.
\newblock Interpreting and executing recipes with a cooking robot.
\newblock In \emph{Experimental Robotics: The 13th International Symposium on Experimental Robotics}, pp.\  481--495. Springer, 2013.

\bibitem[Brohan et~al.(2022)Brohan, Chebotar, Finn, Hausman, Herzog, Ho, Ibarz, Irpan, Jang, Julian, et~al.]{brohan2022can}
Anthony Brohan, Yevgen Chebotar, Chelsea Finn, Karol Hausman, Alexander Herzog, Daniel Ho, Julian Ibarz, Alex Irpan, Eric Jang, Ryan Julian, et~al.
\newblock Do as i can, not as i say: Grounding language in robotic affordances.
\newblock In \emph{6th Annual Conference on Robot Learning}, 2022.

\bibitem[Brown et~al.(2020)Brown, Mann, Ryder, Subbiah, Kaplan, Dhariwal, Neelakantan, Shyam, Sastry, Askell, et~al.]{brown2020language}
Tom Brown, Benjamin Mann, Nick Ryder, Melanie Subbiah, Jared~D Kaplan, Prafulla Dhariwal, Arvind Neelakantan, Pranav Shyam, Girish Sastry, Amanda Askell, et~al.
\newblock Language models are few-shot learners.
\newblock \emph{Advances in neural information processing systems}, 33:\penalty0 1877--1901, 2020.

\bibitem[Chang et~al.(2020)Chang, Huang, Xu, Adeli, Fei-Fei, and Niebles]{chang2020procedure}
Chien-Yi Chang, De-An Huang, Danfei Xu, Ehsan Adeli, Li~Fei-Fei, and Juan~Carlos Niebles.
\newblock Procedure planning in instructional videos.
\newblock In \emph{Computer Vision--ECCV 2020: 16th European Conference, Glasgow, UK, August 23--28, 2020, Proceedings, Part XI}, pp.\  334--350. Springer, 2020.

\bibitem[Dvornik et~al.(2023)Dvornik, Hadji, Zhang, Derpanis, Garg, Wildes, and Jepson]{dvornik2023stepformer}
Nikita Dvornik, Isma Hadji, Ran Zhang, Konstantinos~G Derpanis, Animesh Garg, Richard~P Wildes, and Allan~D Jepson.
\newblock Stepformer: Self-supervised step discovery and localization in instructional videos.
\newblock \emph{arXiv preprint arXiv:2304.13265}, 2023.

\bibitem[Ehsani et~al.(2018)Ehsani, Bagherinezhad, Redmon, Mottaghi, and Farhadi]{ehsani2018let}
Kiana Ehsani, Hessam Bagherinezhad, Joseph Redmon, Roozbeh Mottaghi, and Ali Farhadi.
\newblock Who let the dogs out? modeling dog behavior from visual data.
\newblock In \emph{Proceedings of the IEEE Conference on Computer Vision and Pattern Recognition}, pp.\  4051--4060, 2018.

\bibitem[Jansen(2020)]{jansen2020visually}
Peter Jansen.
\newblock Visually-grounded planning without vision: Language models infer detailed plans from high-level instructions.
\newblock In \emph{Findings of the Association for Computational Linguistics: EMNLP 2020}, pp.\  4412--4417, 2020.

\bibitem[Koller et~al.(2016)Koller, Ney, and Bowden]{koller2016deep}
Oscar Koller, Hermann Ney, and Richard Bowden.
\newblock Deep hand: How to train a cnn on 1 million hand images when your data is continuous and weakly labelled.
\newblock In \emph{Proceedings of the IEEE conference on computer vision and pattern recognition}, pp.\  3793--3802, 2016.

\bibitem[Koller et~al.(2017)Koller, Zargaran, and Ney]{koller2017re}
Oscar Koller, Sepehr Zargaran, and Hermann Ney.
\newblock Re-sign: Re-aligned end-to-end sequence modelling with deep recurrent cnn-hmms.
\newblock In \emph{Proceedings of the IEEE conference on computer vision and pattern recognition}, pp.\  4297--4305, 2017.

\bibitem[Kuehne et~al.(2014)Kuehne, Arslan, and Serre]{breakfast}
Hilde Kuehne, Ali Arslan, and Thomas Serre.
\newblock The language of actions: Recovering the syntax and semantics of goal-directed human activities.
\newblock In \emph{Proceedings of the IEEE conference on computer vision and pattern recognition}, pp.\  780--787, 2014.

\bibitem[Li \& Huang(2023)Li and Huang]{li2023understand}
Mingchen Li and Lifu Huang.
\newblock Understand the dynamic world: An end-to-end knowledge informed framework for open domain entity state tracking.
\newblock \emph{arXiv preprint arXiv:2304.13854}, 2023.

\bibitem[Lin et~al.(2022)Lin, Petroni, Bertasius, Rohrbach, Chang, and Torresani]{lin2022learning}
Xudong Lin, Fabio Petroni, Gedas Bertasius, Marcus Rohrbach, Shih-Fu Chang, and Lorenzo Torresani.
\newblock Learning to recognize procedural activities with distant supervision.
\newblock In \emph{Proceedings of the IEEE/CVF Conference on Computer Vision and Pattern Recognition}, pp.\  13853--13863, 2022.

\bibitem[Menon \& Vondrick(2022)Menon and Vondrick]{menon2022visual}
Sachit Menon and Carl Vondrick.
\newblock Visual classification via description from large language models.
\newblock \emph{arXiv preprint arXiv:2210.07183}, 2022.

\bibitem[Miech et~al.(2019)Miech, Zhukov, Alayrac, Tapaswi, Laptev, and Sivic]{miech2019howto100m}
Antoine Miech, Dimitri Zhukov, Jean-Baptiste Alayrac, Makarand Tapaswi, Ivan Laptev, and Josef Sivic.
\newblock Howto100m: Learning a text-video embedding by watching hundred million narrated video clips.
\newblock In \emph{Proceedings of the IEEE/CVF International Conference on Computer Vision}, pp.\  2630--2640, 2019.

\bibitem[Miech et~al.(2020)Miech, Alayrac, Smaira, Laptev, Sivic, and Zisserman]{miech2020end}
Antoine Miech, Jean-Baptiste Alayrac, Lucas Smaira, Ivan Laptev, Josef Sivic, and Andrew Zisserman.
\newblock End-to-end learning of visual representations from uncurated instructional videos.
\newblock In \emph{Proceedings of the IEEE/CVF Conference on Computer Vision and Pattern Recognition}, pp.\  9879--9889, 2020.

\bibitem[Mishra et~al.(2018)Mishra, Huang, Tandon, Yih, and Clark]{mishra2018tracking}
Bhavana~Dalvi Mishra, Lifu Huang, Niket Tandon, Wen-tau Yih, and Peter Clark.
\newblock Tracking state changes in procedural text: a challenge dataset and models for process paragraph comprehension.
\newblock \emph{arXiv preprint arXiv:1805.06975}, 2018.

\bibitem[Mysore et~al.(2019)Mysore, Jensen, Kim, Huang, Chang, Strubell, Flanigan, McCallum, and Olivetti]{mysore2019materials}
Sheshera Mysore, Zach Jensen, Edward Kim, Kevin Huang, Haw-Shiuan Chang, Emma Strubell, Jeffrey Flanigan, Andrew McCallum, and Elsa Olivetti.
\newblock The materials science procedural text corpus: Annotating materials synthesis procedures with shallow semantic structures.
\newblock \emph{arXiv preprint arXiv:1905.06939}, 2019.

\bibitem[Nishimura et~al.(2021)Nishimura, Hashimoto, Ushiku, Kameko, and Mori]{nishimura2021state}
Taichi Nishimura, Atsushi Hashimoto, Yoshitaka Ushiku, Hirotaka Kameko, and Shinsuke Mori.
\newblock State-aware video procedural captioning.
\newblock In \emph{Proceedings of the 29th ACM International Conference on Multimedia}, pp.\  1766--1774, 2021.

\bibitem[Radford et~al.(2021)Radford, Kim, Hallacy, Ramesh, Goh, Agarwal, Sastry, Askell, Mishkin, Clark, et~al.]{radford2021learning}
Alec Radford, Jong~Wook Kim, Chris Hallacy, Aditya Ramesh, Gabriel Goh, Sandhini Agarwal, Girish Sastry, Amanda Askell, Pamela Mishkin, Jack Clark, et~al.
\newblock Learning transferable visual models from natural language supervision.
\newblock In \emph{International conference on machine learning}, pp.\  8748--8763. PMLR, 2021.

\bibitem[Richard et~al.(2017)Richard, Kuehne, and Gall]{richard2017weakly}
Alexander Richard, Hilde Kuehne, and Juergen Gall.
\newblock Weakly supervised action learning with rnn based fine-to-coarse modeling.
\newblock In \emph{Proceedings of the IEEE conference on Computer Vision and Pattern Recognition}, pp.\  754--763, 2017.

\bibitem[Richard et~al.(2018)Richard, Kuehne, Iqbal, and Gall]{richard2018neuralnetwork}
Alexander Richard, Hilde Kuehne, Ahsan Iqbal, and Juergen Gall.
\newblock Neuralnetwork-viterbi: A framework for weakly supervised video learning.
\newblock In \emph{Proceedings of the IEEE conference on Computer Vision and Pattern Recognition}, pp.\  7386--7395, 2018.

\bibitem[Rohrbach et~al.(2016)Rohrbach, Rohrbach, Regneri, Amin, Andriluka, Pinkal, and Schiele]{rohrbach15ijcv}
Marcus Rohrbach, Anna Rohrbach, Michaela Regneri, Sikandar Amin, Mykhaylo Andriluka, Manfred Pinkal, and Bernt Schiele.
\newblock Recognizing fine-grained and composite activities using hand-centric features and script data.
\newblock 2016.

\bibitem[Saini et~al.(2023)Saini, Wang, Swaminathan, Jayasundara, He, Gupta, and Shrivastava]{saini2023chop}
Nirat Saini, Hanyu Wang, Archana Swaminathan, Vinoj Jayasundara, Bo~He, Kamal Gupta, and Abhinav Shrivastava.
\newblock Chop \& learn: Recognizing and generating object-state compositions.
\newblock In \emph{Proceedings of the IEEE/CVF International Conference on Computer Vision}, pp.\  20247--20258, 2023.

\bibitem[Shirai et~al.(2022)Shirai, Hashimoto, Nishimura, Kameko, Kurita, Ushiku, and Mori]{shirai2022visual}
Keisuke Shirai, Atsushi Hashimoto, Taichi Nishimura, Hirotaka Kameko, Shuhei Kurita, Yoshitaka Ushiku, and Shinsuke Mori.
\newblock Visual recipe flow: A dataset for learning visual state changes of objects with recipe flows.
\newblock In \emph{Proceedings of the 29th International Conference on Computational Linguistics}, pp.\  3570--3577, 2022.

\bibitem[Sou{\v{c}}ek et~al.(2022{\natexlab{a}})Sou{\v{c}}ek, Alayrac, Miech, Laptev, and Sivic]{souvcek2022look}
Tom{\'a}{\v{s}} Sou{\v{c}}ek, Jean-Baptiste Alayrac, Antoine Miech, Ivan Laptev, and Josef Sivic.
\newblock Look for the change: Learning object states and state-modifying actions from untrimmed web videos.
\newblock In \emph{Proceedings of the IEEE/CVF Conference on Computer Vision and Pattern Recognition}, pp.\  13956--13966, 2022{\natexlab{a}}.

\bibitem[Sou{\v{c}}ek et~al.(2022{\natexlab{b}})Sou{\v{c}}ek, Alayrac, Miech, Laptev, and Sivic]{souvcek2022multi}
Tom{\'a}{\v{s}} Sou{\v{c}}ek, Jean-Baptiste Alayrac, Antoine Miech, Ivan Laptev, and Josef Sivic.
\newblock Multi-task learning of object state changes from uncurated videos.
\newblock \emph{arXiv preprint arXiv:2211.13500}, 2022{\natexlab{b}}.

\bibitem[Sou{\v{c}}ek et~al.(2023)Sou{\v{c}}ek, Damen, Wray, Laptev, and Sivic]{souvcek2023genhowto}
Tom{\'a}{\v{s}} Sou{\v{c}}ek, Dima Damen, Michael Wray, Ivan Laptev, and Josef Sivic.
\newblock Genhowto: Learning to generate actions and state transformations from instructional videos.
\newblock \emph{arXiv preprint arXiv:2312.07322}, 2023.

\bibitem[Srinivas et~al.(2018)Srinivas, Jabri, Abbeel, Levine, and Finn]{srinivas2018universal}
Aravind Srinivas, Allan Jabri, Pieter Abbeel, Sergey Levine, and Chelsea Finn.
\newblock Universal planning networks: Learning generalizable representations for visuomotor control.
\newblock In \emph{International Conference on Machine Learning}, pp.\  4732--4741. PMLR, 2018.

\bibitem[Sun et~al.(2022)Sun, Huang, Lu, Liu, Zhou, and Garg]{sun2022plate}
Jiankai Sun, De-An Huang, Bo~Lu, Yun-Hui Liu, Bolei Zhou, and Animesh Garg.
\newblock Plate: Visually-grounded planning with transformers in procedural tasks.
\newblock \emph{IEEE Robotics and Automation Letters}, 7\penalty0 (2):\penalty0 4924--4930, 2022.

\bibitem[Tandon et~al.(2018)Tandon, Mishra, Grus, Yih, Bosselut, and Clark]{tandon2018reasoning}
Niket Tandon, Bhavana~Dalvi Mishra, Joel Grus, Wen-tau Yih, Antoine Bosselut, and Peter Clark.
\newblock Reasoning about actions and state changes by injecting commonsense knowledge.
\newblock \emph{arXiv preprint arXiv:1808.10012}, 2018.

\bibitem[Tandon et~al.(2020)Tandon, Sakaguchi, Mishra, Rajagopal, Clark, Guerquin, Richardson, and Hovy]{tandon2020dataset}
Niket Tandon, Keisuke Sakaguchi, Bhavana~Dalvi Mishra, Dheeraj Rajagopal, Peter Clark, Michal Guerquin, Kyle Richardson, and Eduard Hovy.
\newblock A dataset for tracking entities in open domain procedural text.
\newblock \emph{arXiv preprint arXiv:2011.08092}, 2020.

\bibitem[Tang et~al.(2019)Tang, Ding, Rao, Zheng, Zhang, Zhao, Lu, and Zhou]{tang2019coin}
Yansong Tang, Dajun Ding, Yongming Rao, Yu~Zheng, Danyang Zhang, Lili Zhao, Jiwen Lu, and Jie Zhou.
\newblock Coin: A large-scale dataset for comprehensive instructional video analysis.
\newblock In \emph{Proceedings of the IEEE/CVF Conference on Computer Vision and Pattern Recognition}, pp.\  1207--1216, 2019.

\bibitem[Tellex et~al.(2011)Tellex, Kollar, Dickerson, Walter, Banerjee, Teller, and Roy]{tellex2011understanding}
Stefanie Tellex, Thomas Kollar, Steven Dickerson, Matthew Walter, Ashis Banerjee, Seth Teller, and Nicholas Roy.
\newblock Understanding natural language commands for robotic navigation and mobile manipulation.
\newblock In \emph{Proceedings of the AAAI Conference on Artificial Intelligence}, pp.\  1507--1514, 2011.

\bibitem[Vaswani et~al.(2017)Vaswani, Shazeer, Parmar, Uszkoreit, Jones, Gomez, Kaiser, and Polosukhin]{vaswani2017attention}
Ashish Vaswani, Noam Shazeer, Niki Parmar, Jakob Uszkoreit, Llion Jones, Aidan~N Gomez, {\L}ukasz Kaiser, and Illia Polosukhin.
\newblock Attention is all you need.
\newblock \emph{Advances in neural information processing systems}, 30, 2017.

\bibitem[Viterbi(1967)]{viterbi1967error}
Andrew Viterbi.
\newblock Error bounds for convolutional codes and an asymptotically optimum decoding algorithm.
\newblock \emph{IEEE transactions on Information Theory}, 13\penalty0 (2):\penalty0 260--269, 1967.

\bibitem[Wang et~al.(2023{\natexlab{a}})Wang, Lin, Du, Meng, and Zheng]{wang2023event}
An-Lan Wang, Kun-Yu Lin, Jia-Run Du, Jingke Meng, and Wei-Shi Zheng.
\newblock Event-guided procedure planning from instructional videos with text supervision.
\newblock \emph{arXiv preprint arXiv:2308.08885}, 2023{\natexlab{a}}.

\bibitem[Wang et~al.(2023{\natexlab{b}})Wang, Wu, Guo, and Wang]{wang2023pdpp}
Hanlin Wang, Yilu Wu, Sheng Guo, and Limin Wang.
\newblock Pdpp: Projected diffusion for procedure planning in instructional videos.
\newblock \emph{arXiv preprint arXiv:2303.14676}, 2023{\natexlab{b}}.

\bibitem[Wei et~al.(2022)Wei, Wang, Schuurmans, Bosma, Chi, Le, and Zhou]{wei2022chain}
Jason Wei, Xuezhi Wang, Dale Schuurmans, Maarten Bosma, Ed~Chi, Quoc Le, and Denny Zhou.
\newblock Chain of thought prompting elicits reasoning in large language models.
\newblock \emph{arXiv preprint arXiv:2201.11903}, 2022.

\bibitem[Wu et~al.(2023)Wu, Li, and Ji]{wu2023openpi}
Xueqing Wu, Sha Li, and Heng Ji.
\newblock Openpi-c: A better benchmark and stronger baseline for open-vocabulary state tracking.
\newblock \emph{arXiv preprint arXiv:2306.00887}, 2023.

\bibitem[Xue et~al.(2023)Xue, Ashutosh, and Grauman]{xue2023learning}
Zihui Xue, Kumar Ashutosh, and Kristen Grauman.
\newblock Learning object state changes in videos: An open-world perspective.
\newblock \emph{arXiv preprint arXiv:2312.11782}, 2023.

\bibitem[Yang \& Nyberg(2015)Yang and Nyberg]{yang2015leveraging}
Zi~Yang and Eric Nyberg.
\newblock Leveraging procedural knowledge for task-oriented search.
\newblock In \emph{Proceedings of the 38th International ACM SIGIR Conference on Research and Development in Information Retrieval}, pp.\  513--522, 2015.

\bibitem[Zeng et~al.(2022)Zeng, Attarian, Ichter, Choromanski, Wong, Welker, Tombari, Purohit, Ryoo, Sindhwani, et~al.]{zeng2022socratic}
Andy Zeng, Maria Attarian, Brian Ichter, Krzysztof Choromanski, Adrian Wong, Stefan Welker, Federico Tombari, Aveek Purohit, Michael Ryoo, Vikas Sindhwani, et~al.
\newblock Socratic models: Composing zero-shot multimodal reasoning with language.
\newblock \emph{arXiv preprint arXiv:2204.00598}, 2022.

\bibitem[Zhang et~al.(2020)Zhang, Lyu, and Callison-Burch]{zhang2020reasoning}
Li~Zhang, Qing Lyu, and Chris Callison-Burch.
\newblock Reasoning about goals, steps, and temporal ordering with wikihow.
\newblock \emph{arXiv preprint arXiv:2009.07690}, 2020.

\bibitem[Zhang et~al.(2023)Zhang, Xu, Kommula, Tandon, and Callison-Burch]{zhang2023openpi2}
Li~Zhang, Hainiu Xu, Abhinav Kommula, Niket Tandon, and Chris Callison-Burch.
\newblock Openpi2. 0: An improved dataset for entity tracking in texts.
\newblock \emph{arXiv preprint arXiv:2305.14603}, 2023.

\bibitem[Zhao et~al.(2022)Zhao, Hadji, Dvornik, Derpanis, Wildes, and Jepson]{zhao2022p3iv}
He~Zhao, Isma Hadji, Nikita Dvornik, Konstantinos~G Derpanis, Richard~P Wildes, and Allan~D Jepson.
\newblock P3iv: Probabilistic procedure planning from instructional videos with weak supervision.
\newblock In \emph{Proceedings of the IEEE/CVF Conference on Computer Vision and Pattern Recognition}, pp.\  2938--2948, 2022.

\bibitem[Zhou et~al.(2018)Zhou, Xu, and Corso]{youcook2}
Luowei Zhou, Chenliang Xu, and Jason~J Corso.
\newblock Towards automatic learning of procedures from web instructional videos.
\newblock In \emph{Thirty-Second AAAI Conference on Artificial Intelligence}, 2018.

\bibitem[Zhukov et~al.(2019)Zhukov, Alayrac, Cinbis, Fouhey, Laptev, and Sivic]{zhukov2019cross}
Dimitri Zhukov, Jean-Baptiste Alayrac, Ramazan~Gokberk Cinbis, David Fouhey, Ivan Laptev, and Josef Sivic.
\newblock Cross-task weakly supervised learning from instructional videos.
\newblock In \emph{Proceedings of the IEEE/CVF Conference on Computer Vision and Pattern Recognition}, pp.\  3537--3545, 2019.

\end{thebibliography}
}

\appendix
\section*{Appendix}

\section{Implementation Details}
In Section~\ref{sec:3.3}, we introduced the architecture of our SCHEMA. In the appendix, we further introduce the implementation details of the model, including model architecture and training.

\textbf{Chain-of-Thought Prompting.} Recall that we use a chain-of-thought prompt with examples to trigger the language model, which encourages the language model to generate the descriptions of state changes according to the action description. For example, for the action step ``\texttt{cut banana}'' in the task ``\texttt{make banana ice cream}'' , the form of our prompt is:
\begin{adjustwidth}{1cm}{1cm}
\begin{lstlisting}[breakatwhitespace=true]
First, describe the details of [step] for [goal] with one verb. 
Second, use 3 sentences to describe the status changes of objects before and after [step], avoiding using [verb].

[goal]: Grill steak
[step]: season steak
[verb]: season
Description:
Season steak with salt and pepper
Before:
- The steak is unseasoned.
- The steak has no salt or pepper on it.
- The steak is raw.
After:
- The steak is with salt and pepper.
- The steak has salt and pepper on it.
- The steak is ready to be grilled.

[goal]: Make Kimchi Fried Rice
[step]: add ham
[verb]: add
Description:
Incorporate diced ham into the fried rice
Before:
- The diced ham is separate from the pan.
- The pan contains fried rice.
- The pan has no ham on it.
After:
- The diced ham is blended with the fried rice.
- The ham is on the pan.
- The pan contains ham.

[goal]: Make Banana Ice Cream
[step]: cut banana 
\end{lstlisting}
\end{adjustwidth}

Note that the examples are generated by GPT-3.5 and manually selected. An example output is
\begin{adjustwidth}{1cm}{1cm}
\begin{lstlisting}[breakatwhitespace=true]
[verb]: cut
Description:
Cut banana into small pieces
Before:
- The banana is a whole.
- The banana is uncut.
- The banana is in one piece.
After:
- The banana is in small pieces.
- The banana is cut into pieces.
- The banana is divided into multiple parts.

\end{lstlisting}
\end{adjustwidth}

\revise{\textbf{Transformer Models for State Decoder and Step Decoder.} Our state decoder and step decoder are Transformer-based models. The model consists of two blocks. Each block consists of one self-attention module, one cross-attention module, and a two-layer projection module. The input query is first processed by the self-attention module, then forwarded to the cross-attention module, and processed by the projection module. The cross-attention module takes the external memory to calculate the keys and values. Each self-attention and cross-attention module consists of 32 heads and the hidden layer size is set as 128. The step classifier is a two-layer MLP with hidden size of 128. The dropout ratio is 0.2.}

\revise{\textbf{Training Details.} We train our model with Adam optimizer, an initial learning rate set to $5\times 10^{-3}$ decayed by 0.65 every 40 epochs. The batch size is set as 256. The training process takes 1 hour (500 epochs) on CrossTask and 5.5 hours (400 epochs) on COIN using a single V100 GPU. We will release the code after the paper is accepted. The code will be released under Apache-2.0 license.}

\textbf{Viterbi Algorithm for Inference.} During inference, we follow Zhao \etal~\cite{zhao2022p3iv} and use Viterbi~\cite{viterbi1967error} algorithm to further incorporate the temporal prior into the model. Specifically, Viterbi algorithm is conducted based on a transition matrix and an emission matrix. The transition matrix depicts the probability of state transiting from one state to another state, which is denoted as $A\in\mathbb{R}^{N_a\times N_a}$. We use the action co-occurrence frequencies (\ie, ground-truth procedure annotations in the training set) to obtain the transition matrix as temporal prior knowledge. The emission matrix depicts the probability of each state given observations, which is denoted as $B\in\mathbb{R}^{T\times N_a}$ and obtained by the predicted probability distribution.

\section{Experiments}

\begin{wrapfigure}{r}{0.35\textwidth}
    \vspace{-4mm}
    \includegraphics[width=0.35\textwidth]{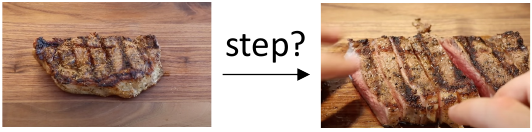}
    \caption{Illustration of state-based step classification.}
    \label{fig:stepcls}
    \vspace{-4mm}
\end{wrapfigure}

\textbf{Baselines}. We follow previous works~\cite{zhao2022p3iv,bi2021procedure,chang2020procedure,sun2022plate} and consider the baseline methods for comparisons. In addition to the baselines in Sec~\ref{sec:eval}, we introduce other baselines as follows:(1) Random. This baseline randomly selects actions from the candidate action set to generate the procedure. (2) Retrieval-based. The baseline retrieves the closest neighbor of the start and goal observations based on the visual feature similarity in the training set. (3) WLTDO~\cite{ehsani2018let}. This baseline used a recurrent neural network (RNN) to predict the action sequence given the visual observations. (4) UAAA~\cite{abu2019uncertainty}. This baseline is an auto-regression model using RNN-HMM architecture. (5) UPN~\cite{srinivas2018universal}. This baseline is a physical-world path planning method. (6) DDN~\cite{chang2020procedure}. This baseline is an auto-regressive model that predicts both the action and state in the same space.

\begin{table}[t]
\centering
\caption{Results on the state-based zero-shot step classification task on CrossTask.}
\scalebox{0.92}{
\begin{tabular}{l|cccc}
\toprule
Prompts & None & ``A photo of'' & ``A video of'' & ``The start/end state of''\\
\hline
Acc (\%) & 14.6 & 14.1 & 16.0 & 13.4 \\
\midrule
Descriptions & DCLIP & Action descriptions & State descriptions & \\
\hline
Acc (\%) & 16.2 & 17.7 & \bf 21.1 & \\
\bottomrule
\end{tabular}
}

\label{tab:stepcls}
\end{table}

\textbf{State Descriptions}. We establish a preliminary task called state-based step classification. As illustrated in Figure~\ref{fig:stepcls}, given two visual state observations, the task is to predict the one action step that leads to the state changes. It can be regarded as a special form of procedure planning with $T=1$. We conduct evaluations under the zero-shot setting using CLIP~\cite{radford2021learning} with different language descriptions, which are listed in Table~\ref{tab:stepcls}. We evaluate both manually-designed prompted captions (first row) and GPT-3.5 generated descriptions introduced in Sec.~\ref{sec:3.2} (second row). Detailed examples can be found in the appendix. As shown in Table~\ref{tab:stepcls}, our state descriptions using chain-of-thought prompting achieves a relative 31.9\% improvement compared to manually-designed prompted captions (21.1 vs. 16.0), while the improvement of the modified baseline 
\revise{DCLIP}~\cite{menon2022visual} is marginal (16.2 vs. 16.0). These results show that our state descriptions are better state descriptors for CLIP.

\begin{table}
\centering
\caption{Comparisons on the NIV dataset.}
\scalebox{0.92}{
\begin{tabular}{l ccc c ccc}
\toprule
& \multicolumn{3}{c}{$T$ = 3} & & \multicolumn{3}{c}{$T$ = 4} \\ 
\cline{2-4} \cline{6-8}
{Models}           & SR$\uparrow$    & mAcc$\uparrow$   & mIoU$\uparrow$  &   & SR$\uparrow$    & mAcc$\uparrow$   & mIoU$\uparrow$ \\ \midrule
Random       &   2.21    &    4.07    &    6.09    &  &   1.12    &    2.73    &    5.84    \\
{DDN}              &   18.41    &    32.54    &    56.56    &  &   15.97    &    27.09    &    53.84    \\
{Ext-GAIL}              &   22.11    &    42.20    &    65.93    &  &   19.91    &    36.31    &    53.84    \\
{P$^3$IV}             &   24.68    &   49.01    &   74.29   &    &   20.14    &   38.36     &   67.29   \\
{EGPP} &   26.05    &   \bf 51.24   &   75.81   &   &   21.37    &  \bf 41.96     &   74.90  \\
\hline
{SCHEMA (Ours)} & \bf 27.93 & 41.64 & \bf 76.77 & & \bf 23.26 & 39.93 & \bf 76.75 \\
\bottomrule
\end{tabular}
}
\label{tab:niv}
\end{table}

\textbf{Results on NIV}. Table~\ref{tab:niv} shows the results on the NIV dataset. Similar to the observations on CrossTask and COIN, our SCHEMA achieves better SR and mIoU and competitive mAcc.

\textbf{Uncertain Modeling}. Uncertain modeling is to produce several procedures by running the model multiple times with different noise input vectors. Although we focus on deterministic modeling, our model can be easily extended to probabilistic modeling by including the noise vectors into queries of state decoder and step decoder. We follow \citet{zhao2022p3iv} to evaluate the performance of uncertain modeling. Evaluation metrics are KL-Div, NLL, Mode Recall (ModeRec), Mode Precision (ModePre). Results are provided in Table~\ref{tab:unc}. \revise{The results of SR, mAcc, and mIoU are shown in Table~\ref{tab:unc2}. As shown in the table, the probabilistic variant underperforms the deterministic variant on procedure planning metrics SR, mAcc, and mIoU. The possible reason is that language descriptions as the supervision of state representations would also decreases the uncertainty and variances of visual observations, which conflicts with noisy vectors in the probabilistic variant that increases the uncertainty and variances.}

\begin{table}[t]
\centering
\caption{The results of uncertain modeling on the CrossTask dataset.}%
\scalebox{0.92}{
\begin{tabular}{l l c c c c}
\toprule %
Metric & Method & $T=3$ & $T=4$ & $T=5$ & $T=6$ \\ %
\midrule %
\multirow{2}*{KL-Div $\downarrow$}
&  Ours - determinstic & 4.03 & 4.31 & 4.49 & 4.65 \\ %
&  Ours - probabilistic & \bf 3.62 & \bf 3.82 & \bf 3.92 & \bf 3.97 \\ %
\midrule %
\multirow{2}*{NLL $\downarrow$}
&  Ours - determinstic & 4.55 & 5.11 & 5.46 & 5.71 \\ %
&  Ours - probabilistic & \bf 4.15 & \bf 4.62 & \bf 4.88 & \bf 5.04 \\ 
\midrule %
\multirow{2}*{ModePrec $\uparrow$}
&  Ours - determinstic & \bf 38.41 & \bf 26.67 & \bf 15.28 & 9.84 \\ %
&  Ours - probabilistic & 38.32 & 26.46 & 14.99 & \bf 9.91 \\
\midrule %
\multirow{2}*{ModeRec $\uparrow$}
&  Ours - determinstic & 25.59 & 13.63 & 6.27 & 3.37
 \\ %
&  Ours - probabilistic & \bf 37.70 & \bf 23.76 & \bf 13.85 & \bf 9.30 \\
\bottomrule %
\end{tabular}
}
\label{tab:unc}
\end{table}

\begin{table}[t]
\centering
\caption{The planning results of probabilistic model on CrossTask.}
\vspace{-3mm}
\scalebox{0.92}{
\begin{tabular}{lccccccccc}
\toprule
& \multicolumn{3}{c}{$T$ = 3} & & \multicolumn{3}{c}{$T$ = 4} & $T$=5 & $T$ = 6 \\ 
\cline{2-4} \cline{6-8}{white}
{Models}           & SR$\uparrow$    & mAcc$\uparrow$   & mIoU$\uparrow$  &   & SR$\uparrow$    & mAcc$\uparrow$   & mIoU$\uparrow$  & SR$\uparrow$ & SR$\uparrow$ \\ \midrule
{Ours - probabilistic} & 29.51 & 57.09 & 77.76 &  &  16.55 & \bf 51.93 & 74.42 & 8.73 & 5.53 \\
{Ours - deterministic} & \bf 31.83 & \bf 57.31 & \bf 78.33  &  & \bf 19.71 & 51.85 & \bf 74.46 & \bf 11.41 & \bf 7.68 \\
\bottomrule
\end{tabular}
}
\label{tab:unc2}
\end{table}

\textbf{Ablations on two-decoder design}. We used two decoders for state prediction and step prediction. An alternative is using one decoder for both the state and action steps. We used the two-decoder design because the intermediate states serving as explicit inputs work better than implicit outputs. Table~\ref{tab:abdecoder} further compares the two designs, showing the effectiveness of the two-decoder design.

\textbf{Ablations on external memory}. We further conduct ablation studies to verify the impact of external memory. The baseline is to learn a memory module with the same size of state descriptions but randomly initialized. Table~\ref{tab:memory} compares different choices of external memory for the Transformer models. As shown in the table, state descriptions works better than random initialized memory, which indicates that the semantic information in state descriptions help with procedure planning.

\revise{Note that we also use state descriptions rather than setp descirptions as external memory for the step decoder. Table~\ref{tab:memory2} shows the comparison between these two variants. As shown in the table, state descriptions as external memory outperforms the variant with step descriptions under all the scenarios. The possible reason is that state descriptions contain more information of object status and serve as supplement to step label supervisions. In addition, as shown in Table~\ref{tab:stepcls}, state descriptions works better than step descriptions for the state-based step classification problem, which indicates that state descriptions are good resources for step recognition.}

\textbf{Ablations on Viterbi}. We applied Viterbi algorithm to add temporal prior. Table~\ref{tab:abviterbi} shows that Viterbi has a positive impact on all the metrics, which indicates that Viterbi successfully includes the temporal ordering knowledge in the training data, i.e., action co-occurrence frequencies. 

\begin{figure}
\centering
\includegraphics[width=1.0\textwidth]{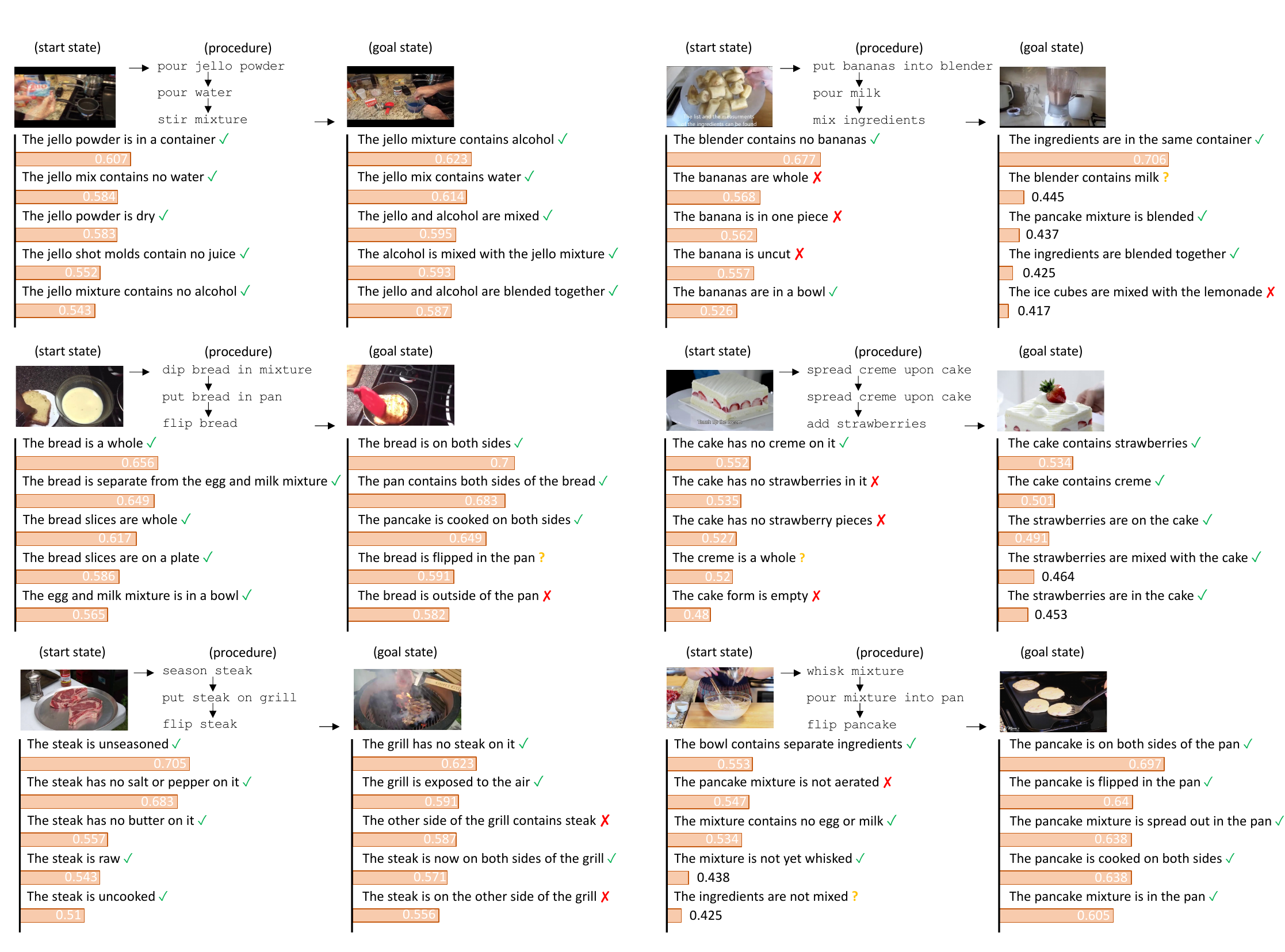}
\caption{More examples of state justifications from our model.}
\label{fig:similarity_full}
\end{figure}

\textbf{Qualitative Results}. Figure~\ref{fig:similarity_full} shows more examples of state justifzication results outputted by our model. These examples show that our model can provide evidence about the visual state, which is more explainable and informative.

\begin{figure}
\centering
\includegraphics[width=1.0\textwidth]{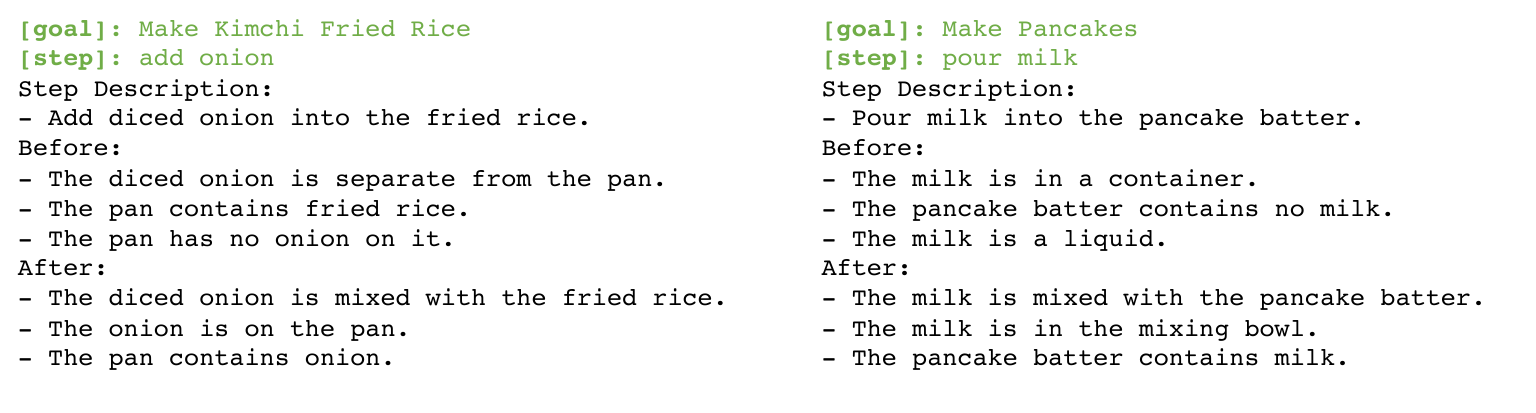}
\caption{Failure case analysis. For each example, the first row is the ground-truth annotation while the second is the prediction.}
\label{fig:fail}
\end{figure}

\revise{\textbf{Failure Case Analysis.} The examples are shown in Figure~\ref{fig:fail}. We grouped them into three cases:}

\revise{(1) Failed understanding of start/end state. As shown in Figure~\ref{fig:fail}(a), the model predicted ``season steak'' as the first step because it didn't recognize that there is pepper on top of the steak, \ie, it failed to understand the start state. As shown in Figure~\ref{fig:fail}(b), the model predicted ``flip steak'' rather than ``put steak on grill'' as the second step. The way to distinguish these two steps is whether the steak has its top side grilled. Although the end state shows that the top side of the steak is raw, the steak is very small to be captured. One future solution is to use high-resolution video frames or object detector to ground the object.}

\revise{(2) Hallucination. As shown in Figure~\ref{fig:fail}(c), the model predicts ``add strawberries to cake'' as the third step. However, the goal is not to make strawberry cake. This failure may be due to the training priors as there are many videos for the task of ``make french strawberry cake''.}

\revise{(3) Reasonable but not matched with ground-truth plans. As shown in Figure~\ref{fig:fail}(d), the generated plan is reasonable, although it doesn't exactly match the ground-truth annotation. This ``failure'' indicates that this task needs a better evaluation protocol for all the reasonable results, which is a general challenge for sequence generation tasks.}

\begin{table}[t!]
\centering
\caption{Ablation studies on decoder design.}
\scalebox{0.92}{
\begin{tabular}{cc c ccc c ccc}
\toprule
& & \multicolumn{3}{c}{$T$ = 3} & & \multicolumn{3}{c}{$T$ = 4} \\
\cline{3-5} \cline{7-9}
 & &
 SR$\uparrow$    & mAcc$\uparrow$   & mIoU$\uparrow$ & & SR$\uparrow$    & mAcc$\uparrow$   & mIoU$\uparrow$ \\
\midrule
 One-decoder & & 30.25 & 56.35 & 77.83 & & 19.01 & 51.15 & 74.13 \\
 Two-decoder &  & \bf 31.83 & \bf 57.31 & \bf 78.33  &  & \bf 19.71 & \bf 51.85 & \bf 74.46 \\
\bottomrule
\end{tabular}
}

\label{tab:abdecoder}
\end{table}

\begin{table}[t!]
\centering
\caption{Ablation studies on external memory.}
\scalebox{0.92}{
\begin{tabular}{cc c ccc c ccc}
\toprule
& & \multicolumn{3}{c}{$T$ = 3} & & \multicolumn{3}{c}{$T$ = 4} \\
\cline{3-5} \cline{7-9}
Memory & &
 SR$\uparrow$    & mAcc$\uparrow$   & mIoU$\uparrow$ & & SR$\uparrow$    & mAcc$\uparrow$   & mIoU$\uparrow$ \\
\midrule
 Random Initialized & & 29.83 & 56.04 & 77.22 & & 19.53 & 50.99 & 73.96 \\
 Two-decoder &  & \bf 31.83 & \bf 57.31 & \bf 78.33  &  & \bf 19.71 & \bf 51.85 & \bf 74.46 \\
\bottomrule
\end{tabular}
}

\label{tab:memory}
\end{table}

\begin{table}[t!]
\centering
\caption{Ablation studies on memory of step classifier.}
\scalebox{0.92}{
\begin{tabular}{cc c ccc c ccccc}
\toprule
& & \multicolumn{3}{c}{$T$ = 3} & & \multicolumn{3}{c}{$T$ = 4} & $T$ = 5 & $T$ = 6 \\
\cline{3-5} \cline{7-9}  
Memory & &
 SR$\uparrow$    & mAcc$\uparrow$   & mIoU$\uparrow$ & & SR$\uparrow$    & mAcc$\uparrow$   & mIoU$\uparrow$ &
 SR$\uparrow$ &
 SR$\uparrow$ \\
\midrule
 Step Descriptions & & 30.00 & 55.89 & 77.61 & & 19.30 & 51.43 & 74.13 & 10.99 & 7.60 \\
 State Descriptions &  & \bf 31.83 & \bf 57.31 & \bf 78.33  &  & \bf 19.71 & \bf 51.85 & \bf 74.46 & \bf 11.41 & \bf 7.68 \\
\bottomrule
\end{tabular}
}

\label{tab:memory2}
\end{table}

\begin{table}[t!]
\centering
\caption{Ablation studies on Viterbi.}
\scalebox{0.92}{
\begin{tabular}{cc c ccc c ccc}
\toprule
& & \multicolumn{3}{c}{$T$ = 3} & & \multicolumn{3}{c}{$T$ = 4} \\
\cline{3-5} \cline{7-9}
 & &
 SR$\uparrow$    & mAcc$\uparrow$   & mIoU$\uparrow$ & & SR$\uparrow$    & mAcc$\uparrow$   & mIoU$\uparrow$ \\
\midrule
 w/o Viterbi & & 27.48 & 56.62 & 68.92 & & 14.85 & 51.46 & 66.47 \\
 w/ Viterbi &  & \bf 31.83 & \bf 57.31 & \bf 78.33  &  & \bf 19.71 & \bf 51.85 & \bf 74.46 \\
\bottomrule
\end{tabular}
}

\label{tab:abviterbi}
\end{table}

\section{Further Discussions}

In this section, we provide further discussions with related works and insights that are not included in the main paper due to page limits.

\revise{\textbf{Limitations.} The limitations of our method are as follows. First, the model may fail to identify state changes if they are not explicitly shown in the video. This is a general challenge for procedure planning models, as the details of objects may be hard to recognize if they are far from the camera. Although we tried to tackle this challenge by associating non-visible state changes with language descriptions, which is learned from the paired vision-text training data, there is no guarantee that the non-visible state changes issue can be totally removed. A potential direction is to generalize the task by taking long-term videos as the input visual observations, so that the model can infer the state changes from video history or ASR narrations. Second, the quality of state descriptions relies on LLMs, which would be a bottleneck of state representation. We will explore more effective and reliable methods to leverage LLMs' commonsense knowledge, or utilize generative vision-language models (VLMs) for state description generation. Third, our approach is under a close-vocabulary setting. We will explore the open-vocabulary setting in the future.}

\revise{\textbf{Differences from other mid-state prediction methods.} Differences from existing mid-state prediction methods: (1) Compared to PlaTe~\cite{sun2022plate} and Ext-GAIL~\cite{bi2021procedure}. First, PlaTe and Ext-GAIL are under the full supervision setting (i.e., visual observations of intermediate states are annotated, shown in Figure 1(b)) while ours is under the weak supervision setting (i.e., mid-state annotations are not available). Second, PlaTe formulates mid-state prediction as $P(s_{t}|s_{t-1},a_{t-1},s_T)$ conditioned on last state $s_{t-1}$, last action $a_{t-1}$, and end state $s_T$, Ext-GAIL formulates mid-state prediction as $P(s_t|s_{t-1},a_{t-1},z_c)$ where  $z_c$ is context information learned from $s_0$ and $s_T$, while we formulate mid-state prediction as $P(s_{1:(T-1)}|s_0,s_T)$ conditioned on start state $s_0$ and end state $s_T$. Third, PlaTe represents mid-states using extremely low-dimensional (actually 4-d) features in their implementation, which is hard to learn and represent mid-state information. The role of their mid-state predictor is questionable.} 

\revise{(2) Compared to methods under the weak supervision settings. Recent works like P3IV~\cite{zhao2022p3iv} consider the weak supervision where mid-states are not annotated, and they didn't predict mid-states in their pipeline. Differently, we leverage LLMs' commonsense knowledge to transform step into state change descriptions. Therefore, our work rethinks the role and representation style of mid-states, and we expect future works to further investigate the state representation.}

\textbf{Comparisons with related works.} Some recent works explored how to extract commonsense information via text descriptions using language models. For example, Socratic Models~\cite{zeng2022socratic} (SMs) use language models in a zero-shot setting to solve downstream multimodal tasks. The similarity between ours and SMs is that we both leveraged LLMs and prompting to obtain commonsense knowledge for other modalities and multimodal tasks, especially video-related tasks and planning tasks. The main differences are: (1) Goal. SMs aim to embrace the heterogeneity of pretrained models through structured Socratic dialogue, while our goal is to represent steps as state changes via LLMs descriptions for procedure planning. (2) Task. SMs focus on zero-shot multimodal tasks without further training, while we focus on procedure learning in instructional videos with weak supervision and further training; (3) Framework. SMs deliver a zero-shot modular framework that composes multimodal models and LLMs and makes sample-specific API calling, while we only used LLMs once to obtain the generic language descriptions and train another planning model.

\textbf{Viterbi alrogithm.} We use Viterbi for post-processing during inference. We use transition matrix in Viterbi to include the temporal ordering knowledge in the training data, i.e., action co-occurrence frequencies. We empirically found that the training priors help with success rate increase. Our emission matrix estimation is a new contribution to deterministic modeling for procedure planning. Instead of sampling 1,500 generated sequences to estimate the emission matrix~\cite{zhao2022p3iv}, we run the feedforwarding only once and use the single predicted probability matrix as the emission matrix, which is simple and more time-efficient. This will be a useful tool for future deterministic modeling works.

\textbf{Hallucination of LLMs.} In the main paper, we observed the hallucination of LLMs using the baseline from \citet{menon2022visual}, \eg, given the action step ``add onion'' and the task goal ``make kimchi fried rice'', the generated description of the state before adding onion is ``the onion was uncut and unchopped'', which is incorrect because the onion should have been cut before being added to the rice. We proposed a chain-of-thought prompting method to first describe more details of the steps and then describe the state changes based on the detailed step descriptions. According to our manual checkup, most of the descriptions are reasonable and match human knowledge. We observed a few failure cases. One common problem is that LLMs may combine two steps as one. For example, for the step ``cut strawberries'' and task "make French strawberry cake", one of the after-state descriptions is ``The strawberries are on the cake'' which is incorrect. The reason is that LLMs mix two steps ``cut strawberries'' and ``add strawberries'' to generate the after-state descriptions. A potential solution is to require the LLMs to (1) distinguish different steps by feeding other steps into the prompt as references, (2) consider the temporal relations between steps.

\textbf{Generalization capabilities.} Our method can handle variations in action steps, object states, and environmental conditions. For variations in action steps, we evaluated our method with variant lengths of action steps, ranging from 3 to 6. According to the experimental results, our method consistently outperforms state-of-the-art models. For object states and environmental conditions, our evaluation benchmark instructional video datasets cover various topics and domains, including different environments like cooking, housework, and car repairing, and different objects like fruits, drinks, and household items. For example, CrossTask covers 133 types of steps in 18 tasks, and COIN covers 778 types of steps in 180 tasks. The evaluation on these two datasets can reflect the models' generality on various object states and environmental conditions.

\textbf{Potential benefit and drawback of mid-states.} The potential benefit of mid-states is the explainability, \eg, extending language-only procedures to multi-modal procedures by adding intermediate visual states. The extension can be realized by (a) retrieving images or video clips from a predefined corpus of images or videos based on feature similarities; (b) generative images via text-to-image generation models (e.g., Stable Diffusion) based on the associated language descriptions. A potential drawback is the impact on the uncertainty modeling, \ie, generating multiple procedures given the same start and goal states. As our motivation is to reduce the high variance of visual observations, language descriptions as the supervision of mid-states would also decrease the uncertainty of sequences. The future direction is to combine mid-states and language descriptions with probabilistic modeling.

\textbf{Noisy visual observations.} Visual scenes of states are diverse and may have low qualities (\ie, incomplete or noisy) in instructional videos. Considering the high variance of visual observations, we proposed to represent visual states using LLMs descriptions and align the visual observations with language descriptions. We expect the discriminative language description to reduce the variance of visual states and add LLM commonsense knowledge for representation learning.

\revise{\textbf{Scaling up to larger dataset.} Different from existing benchmark datasets for procedure planning in instructional videos, larger datasets would have the following requirements: (1) Open-vocabulary setting. Large-scale real-word datasets are often open-vocabulary. For the state description generation process, since LLM is designed for open-vocabulary sequence generation, our generation part can be easily applied to larger datasets. For the step classifier, we can replace the fixed-vocabulary classifier (i.e., two-layer FFN) with a similarity calculation module, i.e., calculating the similarity between output embeddings of step decoder and language step descriptions. (2) Efficient training. Training on large-scale datasets often requires efficient training like distributed computing. Since our state decoder and step decoder are encoder-only Transformer architecture, it is easy to extend the training pipeline to distributed computing. (3) Long-term prediction. As shown in the comparison results with state-of-the-art methods, long-term prediction is still challenging for procedure planning model, but is an essential ability for larger datasets. One potential extension way is to treat our model as an adaptor to transform visual observations into tokens as the inputs to generative model like LLMs or VLMs for long-term sequence generation, which is a popular way to adapt LLMs or VLMs to downstream applications.}

\end{document}